\pdfoutput=1
\documentclass[11pt]{article}
\usepackage[table]{xcolor} 
\usepackage{tikz}
\usepackage{pgfplots}
\pgfplotsset{compat=1.18}
\usetikzlibrary{shapes.geometric, arrows}
\usepackage{amsmath}
\usepackage{amssymb}
\usepackage{amsfonts}
\usepackage{hyperref}
\usepackage{algorithm}
\usepackage[noend]{algpseudocode}

\tikzstyle{block} = [rectangle, rounded corners, minimum width=3cm, minimum height=1cm, text centered, draw=black, fill=blue!30]
\tikzstyle{arrow} = [thick,->,>=stealth]
\tikzstyle{input} = [ellipse, draw, fill=green!30, minimum height=1cm, minimum width=1.5cm, text centered]

\usepackage[utf8]{inputenc} % encode en UTF8

\usepackage[english]{babel} % langue francaise

\usepackage[T1]{fontenc}

\usepackage{caption}

\usepackage{graphicx}

\usepackage[top=2cm, bottom=2cm, left=2cm, right=2cm]{geometry}

\usepackage{amsmath,amsfonts,amssymb}

\usepackage{array}

\usepackage{fancyhdr}

\usepackage{lastpage}

\usepackage[toc,page]{appendix}

\usepackage{algorithm}
\usepackage[noend]{algpseudocode}

\usepackage{ragged2e}
\usepackage{amssymb, bm}
\usepackage{float}
\usepackage{graphicx}

\usepackage{listings}

\usepackage{siunitx}
\usepackage{calrsfs}
\usepackage{titlesec}
\DeclareMathAlphabet{\pazocal}{OMS}{zplm}{m}{n}

\DeclareFixedFont{\ttb}{T1}{txtt}{bx}{n}{12} % for bold
\DeclareFixedFont{\ttm}{T1}{txtt}{m}{n}{12}  % for normal

\definecolor{deepblue}{rgb}{0,0,0.5}
\definecolor{deepred}{rgb}{0.6,0,0}
\definecolor{deepgreen}{rgb}{0,0.5,0}
\definecolor{comment}{rgb}{0.55,0.11,0.46}
\usepackage[utf8]{inputenc}
\usepackage{listings}

\definecolor{codegreen}{rgb}{0,0.6,0}
\definecolor{codegray}{rgb}{0.5,0.5,0.5}
\definecolor{codepurple}{rgb}{0.960, 0.356, 0}
\definecolor{backcolour}{rgb}{0.97, 0.97, 0.97}

\lstdefinestyle{mystyle}{
  backgroundcolor=\color{backcolour},   commentstyle=\color{codegreen},
  keywordstyle=\color{blue},
  numberstyle=\tiny\color{codegray},
  stringstyle=\color{codepurple},
  basicstyle=\ttfamily\tiny,
  breakatwhitespace=false,         
  breaklines=true,                 
  captionpos=b,                    
  keepspaces=true,                 
  numbers=left,                    
  numbersep=5pt,                  
  showspaces=false,                
  showstringspaces=false,
  showtabs=false,                  
  tabsize=2
}
\newcommand\pythonstyle{\lstset{
language=Python,
basicstyle=\ttm,
morekeywords={self},              % Add keywords here
keywordstyle=\ttb\color{deepblue},
emph={MyClass,__init__},          % Custom highlighting
emphstyle=\ttb\color{deepred},    % Custom highlighting style
stringstyle=\color{deepgreen},
frame=tb,                         % Any extra options here
showstringspaces=false,
commentstyle = \ttm\color{comment}
}}

\lstnewenvironment{python}[1][]
{
\pythonstyle
\lstset{#1}
}
{}

\hypersetup{
  colorlinks   = true, %Colours links instead of ugly boxes
  urlcolor     = blue, %Colour for external hyperlinks
  linkcolor    = black, %Colour of internal links
  citecolor   = red %Colour of citations
}

\begin{document}
\renewcommand{\tablename}{Table}
\newcommand\numberthis{\addtocounter{equation}{1}\tag{\theequation}}
\newcommand{\HRule}{\rule{\linewidth}{1mm}}

%%%%%%%%%%%%%%%%%%%%%%%%%%%%%%%%%%%%%%%%%%%%%%%%%%%%%
\def\TP{On Multivariate Financial Time Series Classification}
\def\Redac{Grégory \textsc{Bournassenko\footnote{gregory.bournassenko@etudiants.u-paris2.fr}}}
\def\Date{September 12, 2024}
\def\Encadrant{Mariana \textsc{Rojas Breu}}
% tp.tex — Page de garde style article scientifique

\pagestyle{empty}

\begin{center}
\vspace*{1cm}

{\Large \textbf{\TP}} \\[2em]

{\normalsize \textbf{\Redac}} \\[0.5em]
Université Paris-Panthéon-Assas \\[1.5em]

% \textbf{Supervisor:} \Encadrant \\[0.3em]
% \Date \\[2.5em]
\end{center}

\vspace{2em}

\begin{center}
\begin{minipage}{0.85\textwidth}
\justifying
This article investigates the use of Machine Learning and Deep Learning models in multivariate time series analysis within financial markets. It compares small and big data approaches, focusing on their distinct challenges and the benefits of scaling. Traditional methods such as SVMs are contrasted with modern architectures like ConvTimeNet. The results show the importance of using and understanding Big Data in depth in the analysis and prediction of financial time series.
\end{minipage}
\end{center}

\vfill

%%%%%%%%%%%%%%%%%%%%%%%%%%%%%%%%%%%%%%%%%%%%%%%%%%%%%

%\newpage
%\strut
\newpage
\cleardoublepage
\pagenumbering{roman} 
%\tableofcontents \newpage \strut \cleardoublepage

%%%%%%%%%%%%%%%%%%%%%%%%%%%%%%%%%%%%%%%%%%%%%%%%%%%%%

\pagenumbering{arabic}

\pagestyle{fancy}
\renewcommand{\headrulewidth}{0pt}
\renewcommand\footrulewidth{0.1pt}
\fancyhead[R]{}
\fancyhead[L]{}
\fancyfoot[R]{\thepage/\pageref{LastPage}}
\fancyfoot[L]{\TP}
\fancyfoot[C]{ }

%%%%%%%%%%%%%%%%%%%%%%%%%%%%%%%%%%%%%%%%%%%%%%%%%%%%%
\setcounter{page}{1}

\tableofcontents
\cleardoublepage
\listoffigures
\listoftables
\listofalgorithms
\cleardoublepage

\section{Introduction}
\subsection{Context and Relevance}
\subsubsection{Introduction to time series classification in finance}
The classification of time series in finance is a more or less sophisticated task involving the prediction of future trends in asset prices based on past observations. Unlike traditional regression, which estimates continuous values, classification focuses on categorizing the movement of assets into discrete classes, such as "up", "down", or "neutral" movements.

The formulation of time series classification starts with a time series \( \{x_t\}_{t=1}^N \), where each \( x_t \) represents the observed value of an asset at time \( t \). The classification task aims to assign a label \( y_t \in \{-1, 0, 1\} \) at each time step, where \( -1 \) indicates a downward movement, \( 0 \) a neutral state, and \( 1 \) an upward movement.

The problem is formalized as a mapping:

\[
y_t = f(x_{t-1}, x_{t-2}, \ldots, x_{t-k}; \theta)
\]

where \( f \) is a function parameterized by \( \theta \), and \( k \) is the length of the look-back window. 

\begin{figure}[H]
    \centering
   \includegraphics[width=0.3\linewidth]{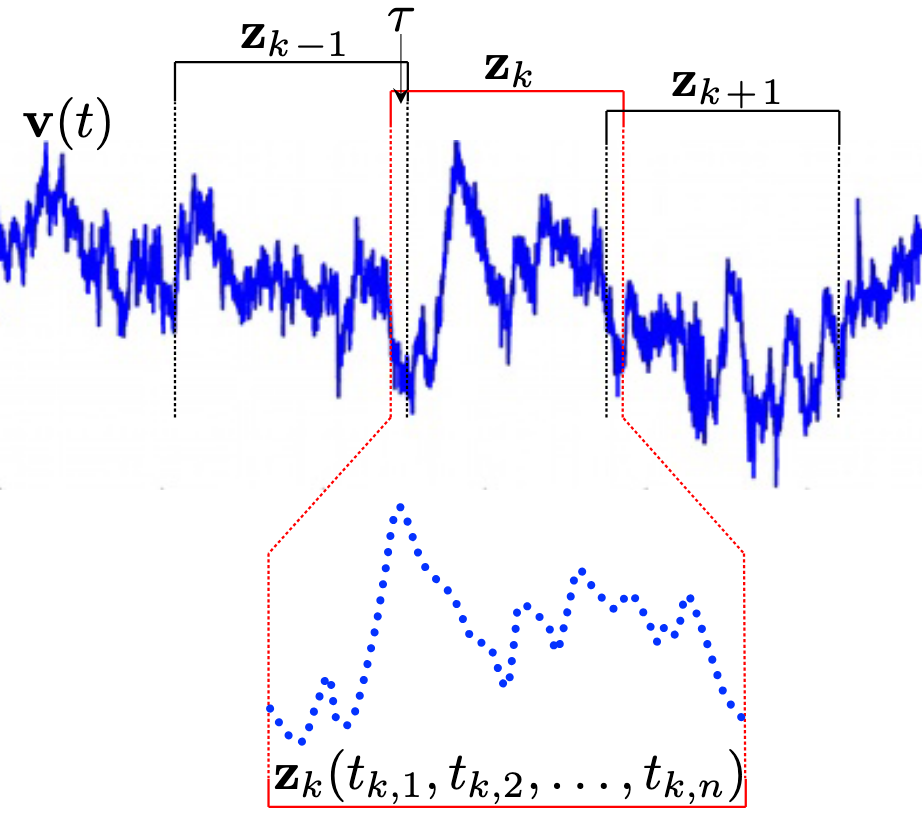}
    \caption{Example of windowing \cite{susto2018time}}
    \label{fig:1}
\end{figure}

A variety of models, such as logistic regression, support vector machines (SVMs), and more recently, deep learning models like recurrent neural networks (RNNs) and long short-term memory (LSTM) networks, have been used to approximate \( f \).

Financial time series, however, pose significant challenges due to their non-stationary nature, meaning that the statistical properties of the series, such as mean and variance, change over time. This makes traditional methods prone to failure, especially when faced with periods of volatility, sudden shocks, or regime shifts. Recent advances in deep learning, particularly in RNNs and LSTMs, allow for better handling of non-stationarity and long-term dependencies by capturing the inherent temporal structure in the data.

A typical recurrent neural network for time series classification can be written as:

\[
h_t = \sigma(W_h h_{t-1} + W_x x_t + b_h)
\]

where \( h_t \) represents the hidden state at time \( t \), and \( \sigma \) is an activation function (commonly \( \tanh \) or ReLU \cite{fukushima1969visual}). The hidden state is then used to predict the class label:

\[
y_t = \arg\max \left( W_y h_t + b_y \right)
\]

where \( W_y \) and \( b_y \) are weights and biases, respectively. This formulation allows the model to learn both short-term patterns (i.e., immediate price movements) and long-term dependencies (i.e., trends over several time steps).

\bigskip
The challenge of non-stationarity in financial time series is often exacerbated by the presence of volatility clustering—periods of high volatility followed by relative calm. This phenomenon is particularly prevalent in stock markets and is modeled using techniques like Generalized Autoregressive Conditional Heteroskedasticity (GARCH) \cite{engle1982introduction}. GARCH models the variance of the error term \( \epsilon_t \) as a function of past squared residuals:

\[
\sigma_t^2 = \alpha_0 + \sum_{i=1}^q \alpha_i \epsilon_{t-i}^2 + \sum_{j=1}^p \beta_j \sigma_{t-j}^2
\]

where \( \alpha_0 \), \( \alpha_i \), and \( \beta_j \) are parameters estimated from data. While GARCH provides a statistical framework for modeling volatility, machine learning techniques like DNNs and LSTMs can capture non-linear relationships and interactions that GARCH models may miss \cite{dixon2017classification}.

\bigskip
Deep neural networks (DNNs) have revolutionized time series classification due to their ability to model complex, non-linear relationships. A DNN consists of multiple layers of neurons, where each layer applies a transformation to the input data. For time series classification, DNNs use historical data points as inputs and pass them through several hidden layers to predict the future state of the series.

The general form of a DNN is:

\[
\hat{y} = \sigma\left( W_L \sigma\left( W_{L-1} \sigma\left( \dots \sigma(W_1 x + b_1) \dots \right) + b_{L-1} \right) + b_L \right)
\]

where \( W_i \) and \( b_i \) are the weights and biases for layer \( i \), and \( \sigma \) is an activation function like the rectified linear unit (ReLU). This architecture allows the model to capture both the short-term fluctuations and long-term trends in financial data \cite{dixon2017classification}.

\begin{figure}[H]
    \centering
   \includegraphics[width=1\linewidth]{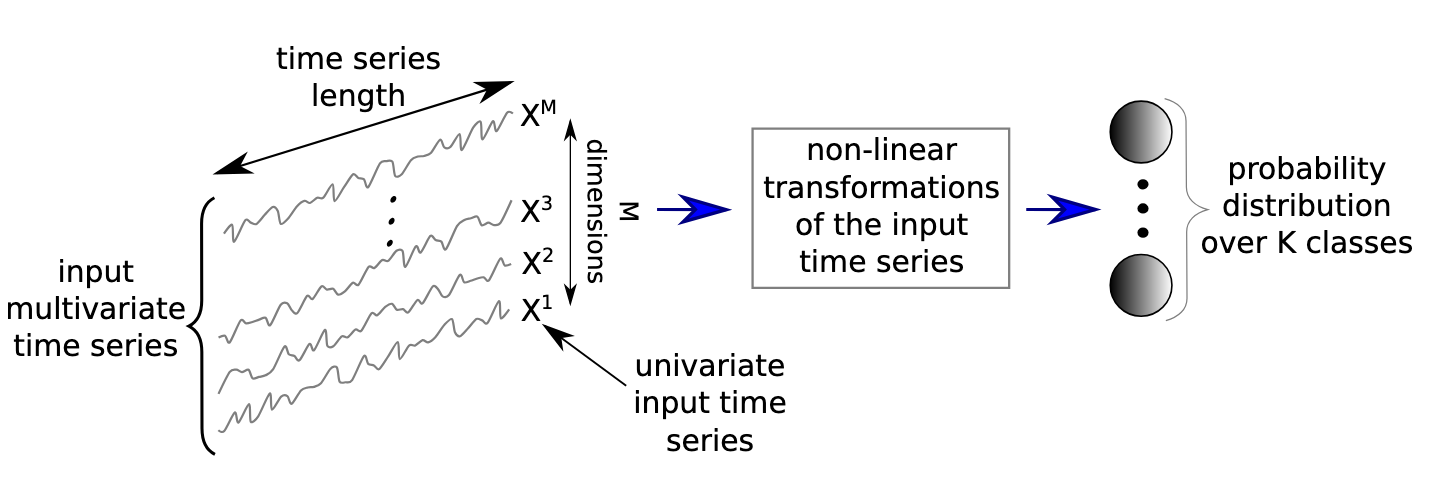}
    \caption{A unified deep learning framework for time series classification \cite{ismail2019deep}}
    \label{fig:1}
\end{figure}

\bigskip
The application of time series classification in finance is widespread, with uses ranging from algorithmic trading to risk management and asset allocation. For instance, Dixon et al. (2017) applied DNNs to classify the direction of future price movements for commodity and FX futures, achieving an accuracy of up to 68\% in some cases \cite{dixon2017classification}. Such models are used to inform trading decisions, allowing firms to develop strategies based on the predicted direction of market movements.

Time series classification has become an essential tool in modern finance. With the growing complexity and volume of financial data, advanced machine learning models, including DNNs, provide powerful tools to predict market trends and inform trading strategies.

\subsubsection{Importance of accurate predictions in the financial market}

Accurate predictions in financial markets are critical due to their direct impact on decision-making processes, including trading strategies, risk management, and portfolio optimization. The ability to forecast market direction allows traders to implement profitable strategies such as long-only trading or mean-reversion strategies. The objective for most trading strategies is to maximize the expected return \( E[R] \) while minimizing the associated risk \( \sigma^2(R) \), where \( R \) denotes the return and \( \sigma \) the standard deviation. For example, a typical long-only strategy can be represented as:

\[
R_t = \log\left(\frac{P_t}{P_{t-1}}\right)
\]

where \( P_t \) is the price at time \( t \). By predicting \( R_t \) accurately, traders can make decisions to buy, sell, or hold based on forecasted price movements.

Modern financial systems rely heavily on algorithmic trading, where predictions are made at high frequency. Models such as DNNs and LSTMs are used to detect patterns in high-frequency time series data, which are often non-linear and contain noise. These models can be optimized using stochastic gradient descent (SGD) and cross-entropy loss functions to improve classification accuracy. The loss function \( \mathcal{L} \) in a multi-class classification setting is defined as:

\[
\mathcal{L}(y, \hat{y}) = -\sum_{i=1}^{C} y_i \log(\hat{y}_i)
\]

where \( C \) is the number of classes (e.g., upward, downward, static), and \( \hat{y}_i \) is the predicted probability for class \( i \) \cite{dixon2017classification}.

\subsubsection{Overview of multivariate time series and its significance}

Multivariate time series (MTS) forecasting is essential in finance because many financial instruments are interdependent. For instance, the movement of stock indices is often correlated with interest rates, exchange rates, commodity prices, etc. In an MTS setting, each time series \( x^i_t \) (for variable \( i \) at time \( t \)) interacts with others, forming a complex network of dependencies. This can be modeled as a system of equations:

\[
X_t = f(X_{t-1}, X_{t-2}, \ldots, X_{t-k}) + \epsilon_t
\]

where \( X_t = [x^1_t, x^2_t, \ldots, x^N_t] \) represents the vector of time series variables at time \( t \), and \( \epsilon_t \) is a noise term. The challenge of MTS is to learn both the temporal dependencies (within a single time series) and the cross-series dependencies (between different time series). For example, Graph Convolution Networks (GCNs) are often used to model the spatial dependencies in MTS, such as the interconnectedness of financial markets \cite{yin2019experimental}.

The importance of MTS is highlighted in applications like portfolio management, where assets are selected based on their joint behavior. In these cases, the goal is often to minimize portfolio risk by considering the covariance matrix \( \Sigma \), where the portfolio variance is given by:

\[
\sigma_p^2 = w^\top \Sigma w
\]

Here, \( w \) is the vector of asset weights, and \( \Sigma \) is the covariance matrix of returns. Understanding the co-movement between time series allows for more informed asset allocation and risk management \cite{wu2020labeling}.

\subsection{Research Objectives and Questions}
\subsubsection{Objective: To explore and compare different methodologies for classifying financial market data using MTS}

The objective of this research is to evaluate and compare some methodologies for multivariate time series (MTS) classification in financial markets. Financial data, being multivariate, contain multiple interrelated time series, such as stock prices, trading volumes, interest rates, etc. which require specialized methods to capture temporal dependencies and correlations between variables. One of the most promising approaches is the use of deep learning techniques, such as Convolutional Neural Networks (CNNs), Long Short-Term Memory networks (LSTMs), and Graph Neural Networks (GNNs) \cite{dixon2017classification, yin2019experimental}. These methods have demonstrated superior performance in time series classification due to their ability to model non-linear relationships, temporal dynamics, and inter-variable dependencies.

MTS can be represented as:

\[
\mathbf{X} = \begin{bmatrix} x_{1,1} & x_{1,2} & \cdots & x_{1,T} \\ x_{2,1} & x_{2,2} & \cdots & x_{2,T} \\ \vdots & \vdots & \ddots & \vdots \\ x_{N,1} & x_{N,2} & \cdots & x_{N,T} \end{bmatrix}
\]

where \( x_{i,t} \) represents the observation of the \(i\)-th variable at time \( t \), and \( N \) is the number of variables. The goal is to predict a label \( y_t \in \{-1, 0, 1\} \), which represents whether the financial market is expected to move downward, remain static, or move upward.

Traditional methods like ARIMA assume linearity and stationarity, which are rarely found in real-world financial data \cite{yin2019experimental}. On the other hand, deep learning models, especially RNNs and LSTMs, can capture long-term dependencies and handle non-linearities. The LSTM model, for example, is defined as:

\[
h_t = \sigma(W_h h_{t-1} + W_x x_t + b_h)
\]

where \( h_t \) is the hidden state at time \( t \), and \( W_h \), \( W_x \), and \( b_h \) are model parameters. LSTMs include gating mechanisms that help mitigate the vanishing gradient problem, making them ideal for financial time series, where long-term dependencies are critical \cite{dixon2017classification}. The output is then passed through a softmax function to predict the label \( y_t \):

\[
y_t = \text{softmax}(W_y h_t + b_y)
\]

\begin{figure}[H]
    \centering
   \includegraphics[width=0.5\linewidth]{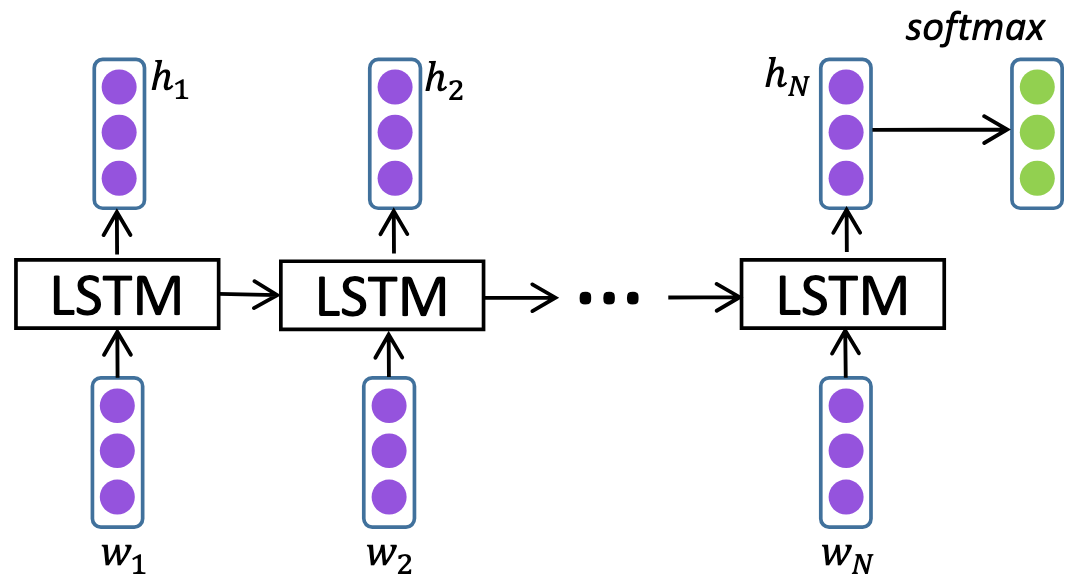}
    \caption{Architecture of a standard LSTM (NLP) \cite{wang2016attention}}
    \label{fig:1}
\end{figure}

\subsubsection{Key questions: How do small data and big data approaches differ? What are the challenges in predicting financial markets with MTS?}

One of the central questions in financial time series classification is how methodologies differ when applied to small versus large datasets. In small datasets, traditional statistical models like ARIMA and simpler machine learning approaches (e.g., Support Vector Machines) may outperform deep learning models due to the latter’s need for large amounts of data for proper generalization \cite{yin2019experimental}. With small data, models are prone to overfitting, and regularization techniques such as \( L_2 \) regularization or early stopping are critical.

\begin{figure}[H]
    \centering
   \includegraphics[width=0.25\linewidth]{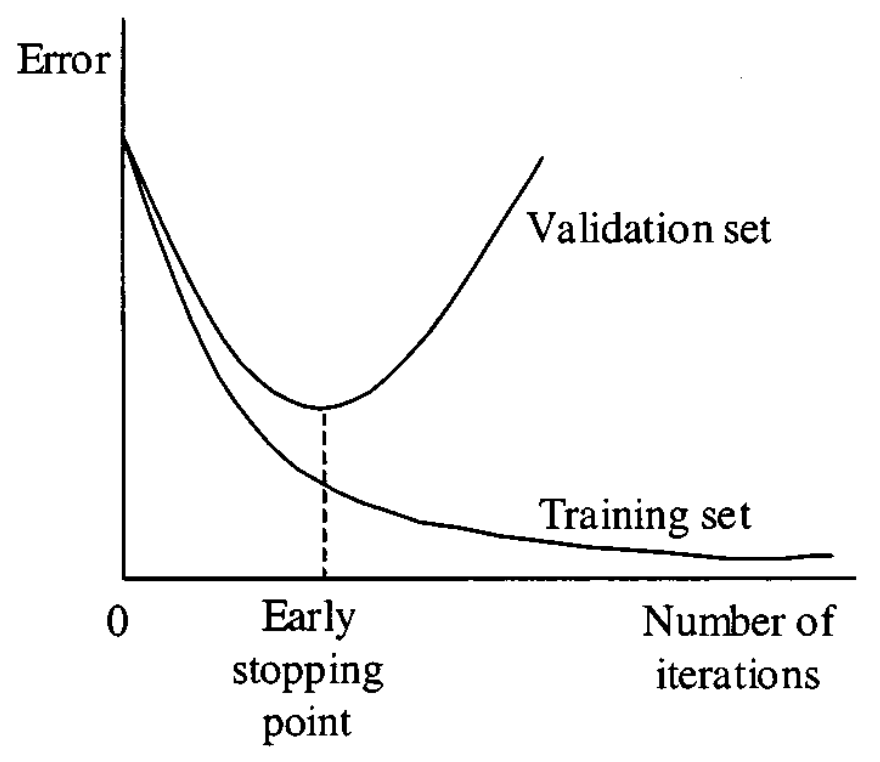}
    \caption{Early stopping based on cross-validation \cite{genccay2001pricing}}
    \label{fig:1}
\end{figure}

The regularized loss function is given by:

\[
\mathcal{L} = \sum_{i=1}^{N} \left( y_i - \hat{y_i} \right)^2 + \lambda \sum_{j=1}^{P} w_j^2
\]

where \( \lambda \) controls the strength of regularization, and \( w_j \) are the model weights.

In contrast, big data approaches such as DNNs and GNNs leverage the sheer volume of data to discover complex patterns and relationships. These models often use techniques like mini-batching and parallelization to handle large-scale data efficiently:

\[
S^{(l)} = \sigma(W^{(l)} X^{(l-1)} + b^{(l)})
\]

where \( S^{(l)} \) is the activation at layer \( l \), and \( W^{(l)} \) and \( b^{(l)} \) are the weights and biases for layer \( l \).

Predicting financial markets with MTS also poses several challenges. The first is the non-stationarity of financial time series—financial markets evolve over time, causing model drift, which necessitates frequent model retraining using walk-forward validation. Second, the presence of noise and volatility in financial data can obscure meaningful patterns, leading to poor predictions if not accounted for. One approach to address this is through feature engineering, such as calculating moving averages and volatility measures like the GARCH model for volatility forecasting:

\[
\sigma_t^2 = \alpha_0 + \sum_{i=1}^{q} \alpha_i \epsilon_{t-i}^2 + \sum_{j=1}^{p} \beta_j \sigma_{t-j}^2
\]

where \( \sigma_t^2 \) is the conditional variance, and \( \epsilon_{t-i}^2 \) are lagged residuals \cite{wu2020labeling}.

The key differences between small data and big data approaches lie in their complexity and data requirements. Big data approaches, particularly deep learning, excel when large datasets are available, while small data methodologies rely on regularization and simpler models to prevent overfitting. Both approaches face the inherent challenges of non-stationarity and volatility, which must be addressed for accurate predictions in financial markets, as we will see later on.

\cleardoublepage

\section{Multivariate Time Series Classification with Small Data}
\subsection{Overview of Small Data in Financial Time Series}

Small data is readily available in financial markets, especially for newly listed assets or niche markets where historical data is limited. An example includes stocks or cryptos that are newly introduced and have only a few weeks or months of price history, as well as using large timeframes over short periods. For instance: AAPL, with a 1-month timeframe from 2010 to 2020 gives 132 periods. Unlike big data, small datasets are often more manageable but present challenges in terms of representativeness and reliability of the predictions. Small datasets may lack enough training data to generalize well in the context of multivariate time series (MTS) classification \cite{goldstein2021big}. However, small data can still be useful when supplemented by carefully chosen features or additional time series to enrich the training process.

\subsubsection{Characteristics of Small Data}

Small data typically refers to datasets that are easily interpretable by humans and manageable using basic computational tools. These datasets are often limited in size and scope, and, as a result, models that operate on small data can be more easily tuned and interpreted. Tools like Support Vector Machines (SVMs) or simple time series methods such as ARIMA are particularly suitable for handling small datasets due to their lower computational complexity compared to deep learning models \cite{yin2019experimental}. The simplicity of these models can be advantageous for small data, where overfitting is a major concern, but they may fall short in capturing complex patterns and interactions.

The function mapping for small data classification can be expressed as:

\[
y_t = f(x_{t-1}, x_{t-2}, \dots, x_{t-k}; \theta) + \epsilon_t
\]

where \( \epsilon_t \) is the noise term, and \( f \) is often a simple linear or non-linear model, like SVM, which can be effective for small datasets but may fail to generalize when data becomes more complex or increases in size.

\subsubsection{Importance of Multivariate vs. Univariate Analysis}

In financial time series, the limited information provided by small datasets requires strategies to supplement the dataset. One such method is to use multivariate time series (MTS) analysis, where several related time series are considered together, capturing the interdependencies between variables \cite{wu2020labeling}. In MTS classification, the goal is to use additional series to enrich the data and allow for more robust predictions. This can be crucial when the training data is small, as relying solely on one time series could lead to overfitting.

For example, suppose \( X_t \) represents multiple variables like stock price, trading volume, and interest rates. The multivariate structure is represented by:

\[
\mathbf{X}_t = \{x_{1,t}, x_{2,t}, \dots, x_{N,t}\}
\]

where \( N \) is the number of time series variables. These variables can provide additional context and improve classification outcomes by introducing inter-variable relationships.

The classification task remains the same:

\[
y_t = \arg \max p(y | \mathbf{X}_{t-1}, \mathbf{X}_{t-2}, \dots, \mathbf{X}_{t-k})
\]

However, by incorporating multiple time series, the model benefits from a richer data representation. Nevertheless, overfitting remains a challenge with small data, as models may fit the noise rather than the underlying pattern, leading to poor generalization when applied to new or live data \cite{goldstein2021big}.

\subsubsection{Small Data Example}

Consider the cryptocurrency \href{https://tokeninsight.com/en/coins/strike-strike/overview}{Strike}, which has been listed on June 4\textsuperscript{th}, 2024, resulting in a total of 107 observations for daily intervals.

\begin{figure}[H]
    \centering
   \includegraphics[width=1\linewidth]{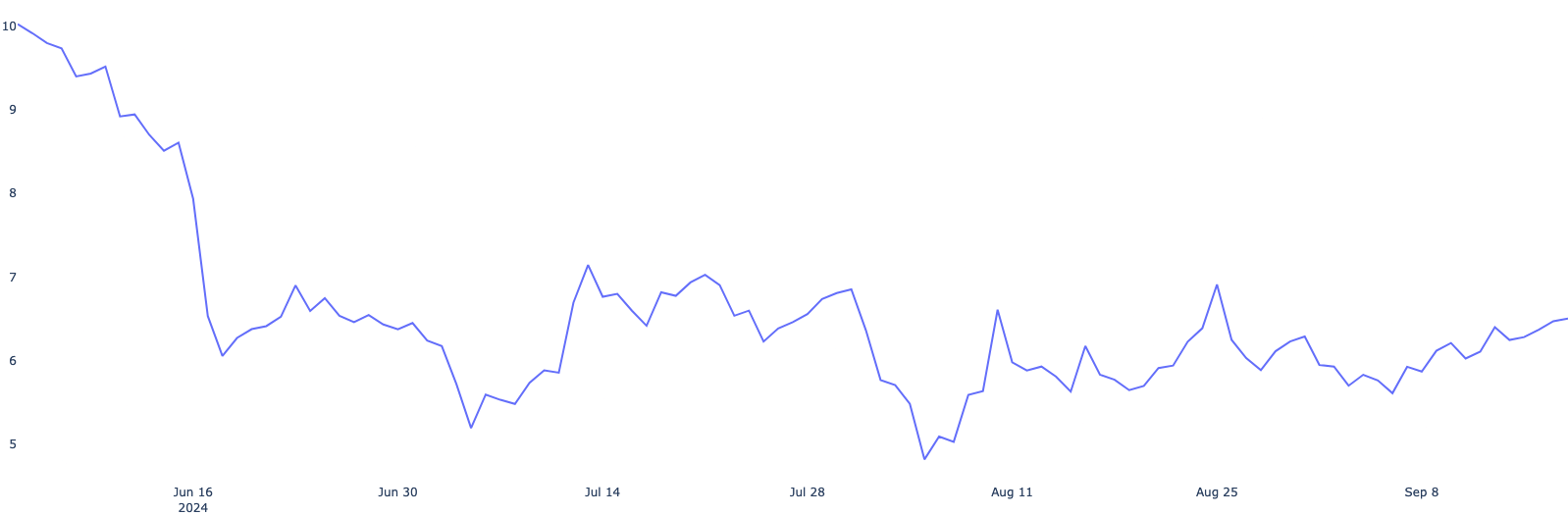}
    \caption{Strike daily closing price (2024-06-04 to 2024-09-18)}
    \label{fig:1}
\end{figure}

% Insert graph of XYZCoin time series data here

We apply an XGBoost model, whose code is in [\textbf{Appendix \ref{appendix:strikeXGBoost}}], to classify future price movements based on this small dataset. After training, we note that the model outperformed the cumulative returns of the market and the random strategy for most of the backtest period.

\begin{figure}[H]
    \centering
   \includegraphics[width=1\linewidth]{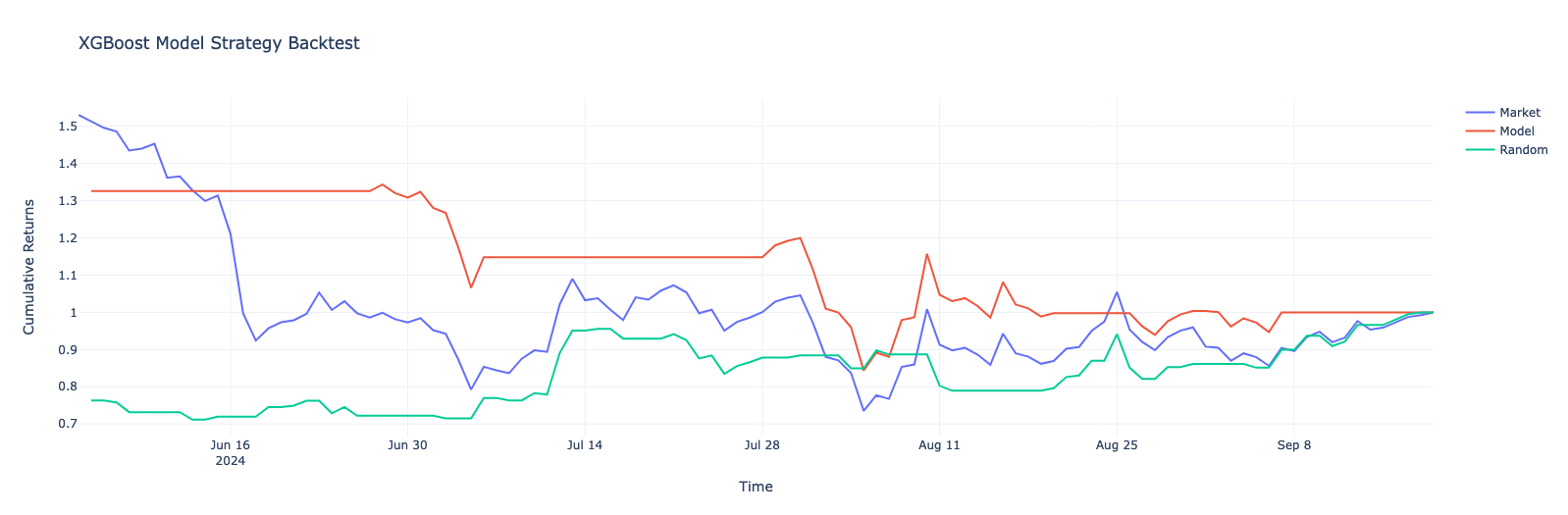}
    \caption{XGBoost vs Model vs Random (Strike)}
    \label{fig:1}
\end{figure}

While this result may appear promising, it is often the result of overfitting, where the model has learned to memorize the small dataset rather than generalizing to unseen data.

To highlight the limitations of small data, we train the same XGBoost model to a more established cryptocurrency, such as Bitcoin, and observe a significantly weaker performance over time. 

\begin{figure}[H]
    \centering
   \includegraphics[width=1\linewidth]{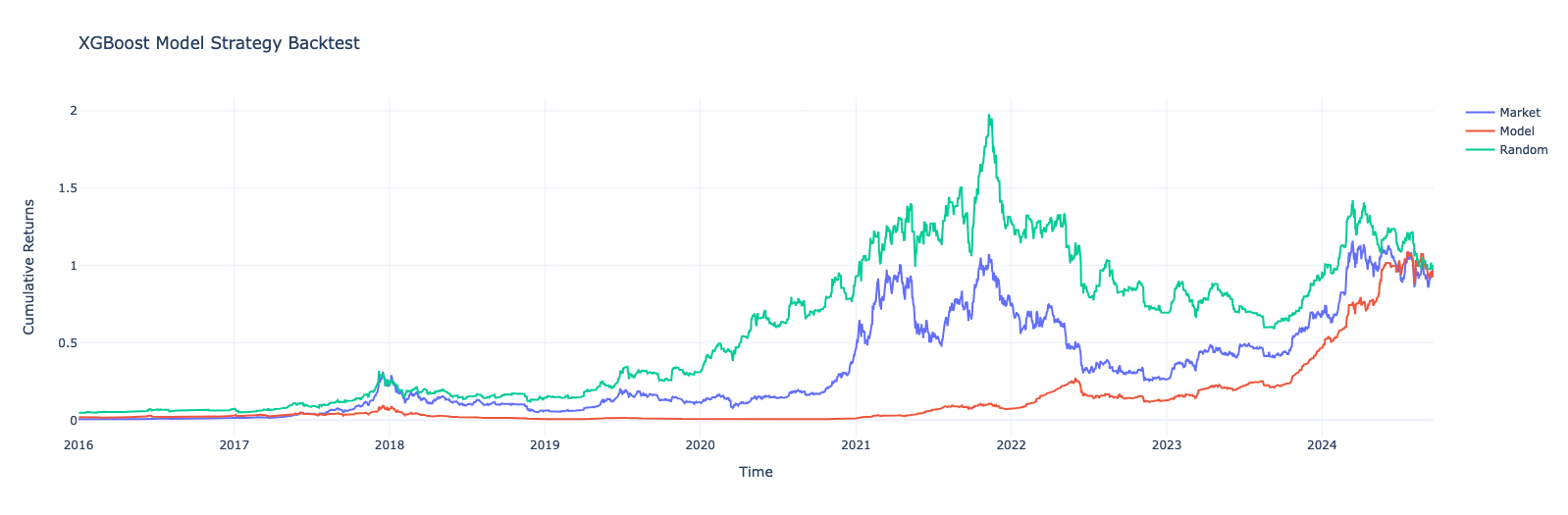}
    \caption{XGBoost vs Model vs Random (Bitcoin)}
    \label{fig:1}
\end{figure}

This is worse than a random trading strategy, demonstrating that models trained on small data can fail to generalize when exposed to larger, more complex datasets \cite{goldstein2021big}.

This experiment illustrates the importance of multivariate analysis and larger datasets in financial time series classification. As models need to learn complex temporal and cross-variable relationships, relying solely on small data leads to results that may not be robust in real-world applications.

\subsection{Machine Learning Approaches for Small Data}
\subsubsection{Traditional Machine Learning Techniques}
There are numerous papers using traditional, non-deep learning techniques for time series classification, but for this section, we will focus on Support Vector Machines (SVM). SVM, introduced by Vapnik and Cortes \cite{cortes1995support}, is particularly interesting for financial data classification because it efficiently handles the problem of small datasets with high-dimensional features.

One advantage of SVM in financial markets is that stock returns often follow a distribution close to a normal distribution, and this can be effectively captured using the Radial Basis Function (RBF) kernel. The RBF kernel introduces non-linearity into the decision surface by transforming the input space into a high-dimensional feature space, where a linear separation becomes possible.

\begin{figure}[H]
    \centering
   \includegraphics[width=0.6\linewidth]{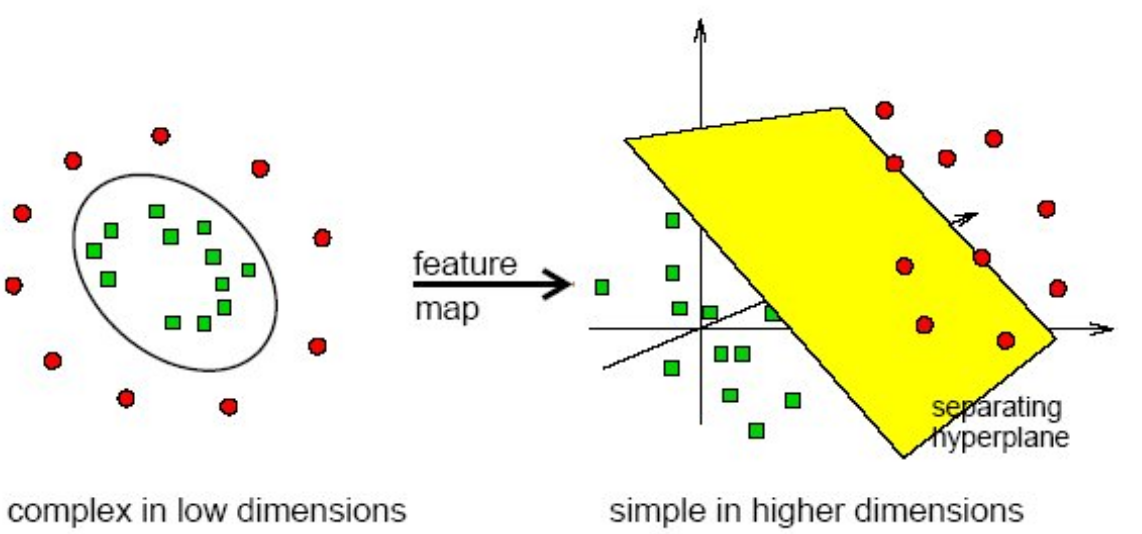}
    \caption{Kernel trick intuition \cite{kwak2013nonlinear}}
    \label{fig:1}
\end{figure}

For financial data, this is particularly useful when modeling returns, volatility, or other features that exhibit complex, non-linear relationships.

The SVM algorithm finds a hyperplane that maximizes the margin between two classes, making it an effective classifier even when data is sparse or limited. The decision function for SVM is:

\[
f(x) = \text{sign}(w \cdot x + b)
\]

where \( w \) is the weight vector, \( x \) is the input feature vector, and \( b \) is the bias term. The goal is to minimize the norm \( ||w|| \), which geometrically corresponds to maximizing the margin between the support vectors (points closest to the separating hyperplane).

\bigskip
Let the training set be defined as \( \{(x_i, y_i)\}_{i=1}^N \), where \( x_i \in \mathbb{R}^n \) are input vectors, and \( y_i \in \{-1, 1\} \) are the corresponding class labels. The primal optimization problem for the SVM classifier can be written as:

\[
\min_{w, b} \frac{1}{2} ||w||^2 \quad \text{subject to} \quad y_i (w \cdot x_i + b) \geq 1, \quad \forall i
\]

This objective function maximizes the margin between the classes. If the data is not linearly separable, the formulation is extended to allow some misclassifications using slack variables \( \xi_i \):

\[
\min_{w, b, \xi} \frac{1}{2} ||w||^2 + C \sum_{i=1}^N \xi_i \quad \text{subject to} \quad y_i (w \cdot x_i + b) \geq 1 - \xi_i, \quad \xi_i \geq 0
\]

where \( C \) is a regularization parameter that controls the trade-off between maximizing the margin and minimizing the classification error.

To solve this problem in a high-dimensional feature space, we use the kernel trick. The dual form of the SVM can be written as:

\[
\max_{\alpha} \sum_{i=1}^N \alpha_i - \frac{1}{2} \sum_{i=1}^N \sum_{j=1}^N \alpha_i \alpha_j y_i y_j K(x_i, x_j)
\]

where \( \alpha_i \) are the Lagrange multipliers, and \( K(x_i, x_j) \) is the kernel function, such as the RBF kernel:

\[
K(x_i, x_j) = \exp \left(-\gamma ||x_i - x_j||^2 \right)
\]

Once the optimization is complete, the decision function becomes:

\[
f(x) = \text{sign}\left( \sum_{i=1}^N \alpha_i y_i K(x_i, x) + b \right)
\]

\begin{figure}[H]
    \centering
   \includegraphics[width=0.4\linewidth]{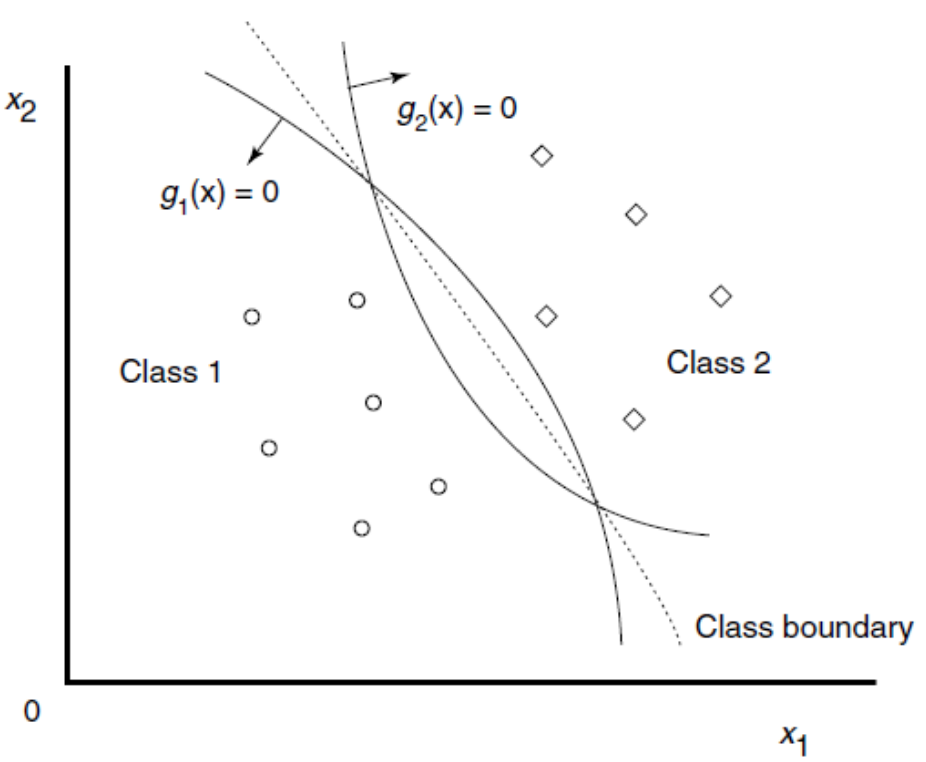}
    \caption{Example of SVM decision function in two dimensional field \cite{abe2010feature}}
    \label{fig:1}
\end{figure}

\begin{algorithm}
\caption{Support Vector Machine (Small Data)}\label{alg:svm}
\begin{algorithmic}[1]
\State Initialize $\alpha_i = 0$ for all $i$
\While{not converged}
    \For{$i = 1$ to $N$}
        \State Solve:
        \[
        \max_{\alpha} \sum_{i=1}^N \alpha_i - \frac{1}{2} \sum_{i=1}^N \sum_{j=1}^N \alpha_i \alpha_j y_i y_j K(x_i, x_j)
        \]
        \State Subject to:
        \[
        0 \leq \alpha_i \leq C, \quad \sum_{i=1}^N \alpha_i y_i = 0
        \]
        \State Update $w = \sum_{i=1}^N \alpha_i y_i x_i$
        \State Compute bias $b$
    \EndFor
\EndWhile
\State Return decision function $f(x) = \text{sign}\left( \sum_{i=1}^N \alpha_i y_i K(x_i, x) + b \right)$
\end{algorithmic}
\end{algorithm}

SVM, due to its reliance on a subset of the training data (support vectors), is particularly effective for small datasets, making it ideal for applications in financial time series with limited historical data. However, it can suffer from overfitting when the dataset is extremely small, and the choice of kernel significantly impacts the performance \cite{hearst1998support}.

\subsubsection{Deep Learning Techniques}

Recurrent Neural Networks (RNNs) are a class of neural networks that are well-suited for sequential data, such as time series. The key feature of RNNs is their internal state (memory), which allows them to capture temporal dependencies between input data points. In financial time series classification, this property enables RNNs to model the sequential nature of market prices, volumes, and other time-related financial indicators.

\bigskip
The RNN model can be mathematically defined as follows:

\[
h_t = \sigma(W_{hh} h_{t-1} + W_{xh} x_t + b_h)
\]
\[
y_t = W_{hy} h_t + b_y
\]

where \( h_t \) is the hidden state at time \( t \), \( x_t \) is the input vector, \( W_{hh} \), \( W_{xh} \), and \( W_{hy} \) are weight matrices, \( b_h \) and \( b_y \) are bias terms, and \( \sigma \) is an activation function such as \( \tanh \). The hidden state \( h_t \) is updated iteratively, and the output \( y_t \) is computed using the current hidden state.

\bigskip
While RNNs are powerful, they suffer from the problem of vanishing gradients, which makes it difficult to capture long-term dependencies. To address this, Long Short-Term Memory (LSTM) networks were introduced.

\begin{figure}[H]
    \centering
   \includegraphics[width=1\linewidth]{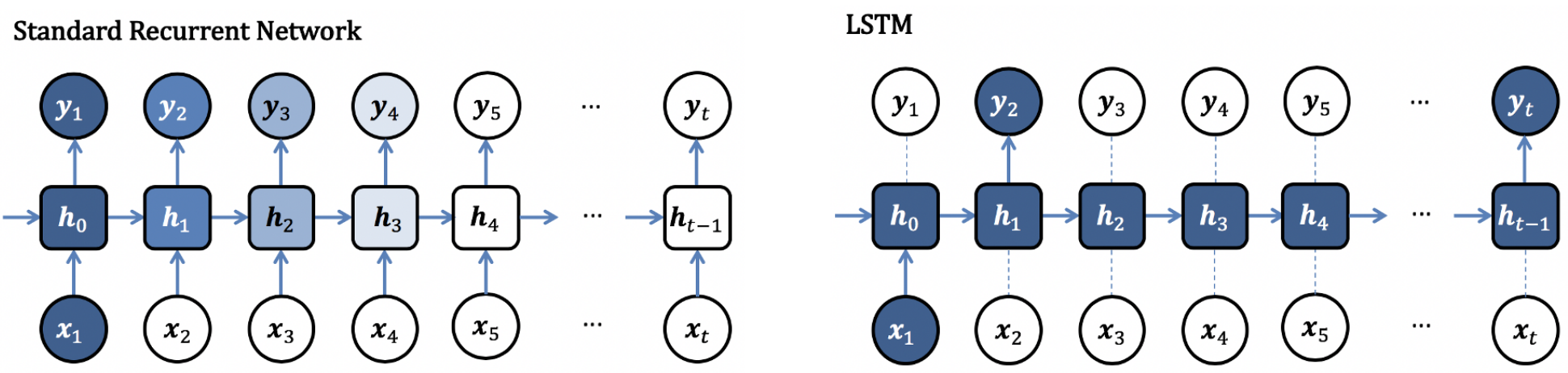}
    \caption{The first figure illustrates a standard recurrent neural network (RNN) and highlights the issue of the vanishing gradient, which leads to a loss of contextual information. In contrast, the second figure demonstrates how LSTM networks retain both information and context. \cite{vanishing}}
    \label{fig:1}
\end{figure}

LSTM networks are a special type of RNN that can learn long-term dependencies in sequential data. They achieve this by incorporating a memory cell and a set of gates—input, forget, and output gates—that control the flow of information into and out of the memory cell.

The LSTM equations are as follows:

\[
f_t = \sigma(W_f \cdot [h_{t-1}, x_t] + b_f)
\]
\[
i_t = \sigma(W_i \cdot [h_{t-1}, x_t] + b_i)
\]
\[
\tilde{C}_t = \tanh(W_C \cdot [h_{t-1}, x_t] + b_C)
\]
\[
C_t = f_t * C_{t-1} + i_t * \tilde{C}_t
\]
\[
o_t = \sigma(W_o \cdot [h_{t-1}, x_t] + b_o)
\]
\[
h_t = o_t * \tanh(C_t)
\]

where \( f_t \), \( i_t \), and \( o_t \) are the forget, input, and output gates, respectively. The cell state \( C_t \) carries the long-term memory, while \( h_t \) represents the hidden state.

\bigskip
LSTMs are particularly effective in financial time series because they can model both short-term fluctuations and long-term trends, making them suitable for predicting stock prices, interest rates, and other financial metrics over time \cite{rumelhart1986learning, hochreiter1997long}. When dealing with small financial datasets, avoiding overfitting is critical. One common approach is to use data augmentation techniques, which generate new training samples by applying transformations to the original dataset. For time series data, augmentation methods include time warping, jittering (adding noise), and window slicing \cite{fons2020evaluating}. These techniques can help increase the diversity of the training set and improve model generalization.

\bigskip
Transfer learning is another powerful technique, especially when there is insufficient data. By pre-training a model on a related large dataset and then fine-tuning it on the small target dataset, transfer learning can significantly improve performance \cite{he2019transfer}. The general approach involves first learning general features from a source dataset and then transferring those features to the target dataset:

\[
h_t^{(source)} = \sigma(W_{source} \cdot x_t^{(source)})
\]
\[
h_t^{(target)} = \sigma(W_{target} \cdot h_t^{(source)})
\]

Here, \( h_t^{(source)} \) represents the learned representation from the source dataset, and \( h_t^{(target)} \) is fine-tuned for the target task.

\begin{figure}[H]
\centering
\begin{minipage}{.5\textwidth}
    \centering
   \includegraphics[width=0.6\linewidth]{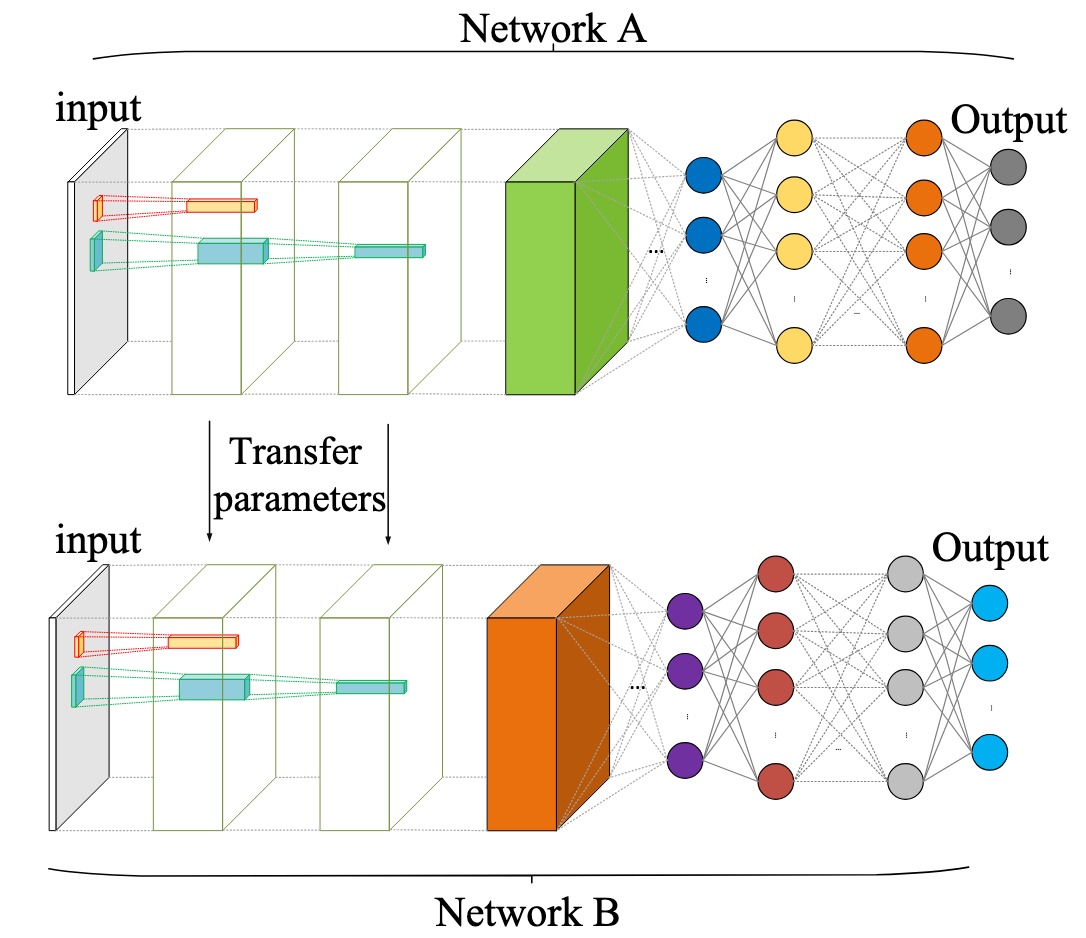}
    \caption{Transfer learning intuition \cite{transferfigure}}
    \label{fig:1}
\end{minipage}%
\begin{minipage}{.5\textwidth}
These strategies, combined with careful hyperparameter tuning and regularization techniques such as dropout, can prevent overfitting and improve the robustness of the model in small financial datasets \cite{fons2020evaluating, he2019transfer}.
\end{minipage}
\end{figure}

\begin{algorithm}
\caption{LSTM Training with Transfer Learning}\label{alg:lstm}
\begin{algorithmic}[1]
\State \textbf{Pre-training on Source Dataset:}
\State Initialize LSTM parameters: \( W_f, W_i, W_C, W_o, b_f, b_i, b_C, b_o \)
\For{each time step \( t \)}
    \State Compute forget gate:
    \[
    f_t = \sigma(W_f \cdot [h_{t-1}, x_t] + b_f)
    \]
    \State Compute input gate:
    \[
    i_t = \sigma(W_i \cdot [h_{t-1}, x_t] + b_i)
    \]
    \State Compute cell update:
    \[
    \tilde{C}_t = \tanh(W_C \cdot [h_{t-1}, x_t] + b_C)
    \]
    \State Update cell state:
    \[
    C_t = f_t \odot C_{t-1} + i_t \odot \tilde{C}_t
    \]
    \State Compute output gate:
    \[
    o_t = \sigma(W_o \cdot [h_{t-1}, x_t] + b_o)
    \]
    \State Compute hidden state:
    \[
    h_t = o_t \odot \tanh(C_t)
    \]
\EndFor
\State Freeze weights \( W_f, W_i, W_C, W_o \)

\State \textbf{Fine-tuning on Target Dataset:}
\State Update LSTM parameters \( W_{f}^{\text{target}}, W_{i}^{\text{target}}, W_{C}^{\text{target}}, W_{o}^{\text{target}}, b_f^{\text{target}}, b_i^{\text{target}}, b_C^{\text{target}}, b_o^{\text{target}} \)
\For{each time step \( t \)}
    \State Repeat computations using target dataset:
    \[
    f_t^{\text{target}} = \sigma(W_f^{\text{target}} \cdot [h_{t-1}^{\text{target}}, x_t^{\text{target}}] + b_f^{\text{target}})
    \]
    \[
    i_t^{\text{target}} = \sigma(W_i^{\text{target}} \cdot [h_{t-1}^{\text{target}}, x_t^{\text{target}}] + b_i^{\text{target}})
    \]
    \[
    \tilde{C}_t^{\text{target}} = \tanh(W_C^{\text{target}} \cdot [h_{t-1}^{\text{target}}, x_t^{\text{target}}] + b_C^{\text{target}})
    \]
    \[
    C_t^{\text{target}} = f_t^{\text{target}} \odot C_{t-1}^{\text{target}} + i_t^{\text{target}} \odot \tilde{C}_t^{\text{target}}
    \]
    \[
    o_t^{\text{target}} = \sigma(W_o^{\text{target}} \cdot [h_{t-1}^{\text{target}}, x_t^{\text{target}}] + b_o^{\text{target}})
    \]
    \[
    h_t^{\text{target}} = o_t^{\text{target}} \odot \tanh(C_t^{\text{target}})
    \]
\EndFor
\State Return the fine-tuned model for target dataset predictions
\end{algorithmic}
\end{algorithm}

\subsection{Evaluation and Application}
\subsubsection{Evaluating Classification Models}

To evaluate the performance of classification models for multivariate financial time series, several metrics are commonly used. In this section, we focus on four key methods: confusion matrix, classification report, Sharpe ratio, and backtesting.

\paragraph{Confusion Matrix}

The confusion matrix is a table that provides a summary of the predictions made by the classification model compared to the actual outcomes. It consists of four key elements for binary classification: True Positives (TP), True Negatives (TN), False Positives (FP), and False Negatives (FN).

The confusion matrix is defined as:

\[
\text{Confusion Matrix} = \begin{pmatrix}
    \text{TP} & \text{FP} \\
    \text{FN} & \text{TN}
\end{pmatrix}
\]

For a multivariate time series classification problem, this matrix can be extended to multiple classes. The accuracy, precision, recall, and F1-score can be derived from this matrix:

\[
\text{Accuracy} = \frac{\text{TP} + \text{TN}}{\text{TP} + \text{TN} + \text{FP} + \text{FN}}
\]
\[
\text{Precision} = \frac{\text{TP}}{\text{TP} + \text{FP}}
\]
\[
\text{Recall} = \frac{\text{TP}}{\text{TP} + \text{FN}}
\]
\[
\text{F1-Score} = 2 \times \frac{\text{Precision} \times \text{Recall}}{\text{Precision} + \text{Recall}}
\]

For example, if the confusion matrix of a stock price movement classification (up, down, stay) is:

\[
\begin{pmatrix}
    30 & 5 & 10 \\
    7 & 40 & 3 \\
    8 & 2 & 45
\end{pmatrix}
\]

it can be used to calculate the classification accuracy and other metrics.

\paragraph{Classification Report}

A classification report provides a detailed analysis of the performance of a classifier, summarizing precision, recall, and F1-score for each class in a multiclass problem. These values are computed using the confusion matrix as follows:

\[
\text{Precision (class i)} = \frac{\text{True Positives for class i}}{\text{True Positives for class i} + \text{False Positives for class i}}
\]
\[
\text{Recall (class i)} = \frac{\text{True Positives for class i}}{\text{True Positives for class i} + \text{False Negatives for class i}}
\]
\[
\text{F1-Score (class i)} = 2 \times \frac{\text{Precision (class i)} \times \text{Recall (class i)}}{\text{Precision (class i)} + \text{Recall (class i)}}
\]

In financial time series, a classification report provides insights into how well a model performs for different types of market movements.

\paragraph{Sharpe Ratio}

The Sharpe ratio is a measure used to evaluate the risk-adjusted return of a trading strategy. It compares the expected return of the strategy to the risk (volatility) involved. The formula for the Sharpe ratio is:

\[
\text{Sharpe Ratio} = \frac{\mathbb{E}[R_t - R_f]}{\sigma_R}
\]

where \( \mathbb{E}[R_t] \) is the expected return of the strategy, \( R_f \) is the risk-free rate, and \( \sigma_R \) is the standard deviation (volatility) of returns.

For a strategy applied to stock returns, if the model predicts future prices or price changes, the Sharpe ratio helps in assessing whether the predictions provide a favorable risk-return balance.

\paragraph{Backtesting}

Backtesting involves applying the model's predictions to historical financial data to simulate how it would have performed in real-market scenarios. The performance of the strategy is then evaluated by comparing the predicted trades to actual market outcomes.

For example, let \( P_t \) be the predicted price at time \( t \), and \( S_t \) the actual price. The profit \( \Pi_t \) from a trade can be defined as:

\[
\Pi_t = (P_t - S_t) \cdot Q_t
\]

where \( Q_t \) is the quantity traded. The total profit or loss over the backtesting period is:

\[
\text{Total Profit} = \sum_{t=1}^{T} \Pi_t
\]

Backtesting allows the evaluation of a model’s real-world applicability by determining how well it predicts market movements and generates returns.

\subsubsection{Application on EUR/USD}
We aim to classify future price movements using various machine learning models on a multivariate time series dataset. Here is a description of the process followed in the implementation, whose code is in [\textbf{Appendix \ref{appendix:smallDataApplication}}].

\bigskip
Given a time series \( P(t) \) of (closing) prices, the next period return \( R(t+1) \) is calculated as the percentage change in the closing price from time \( t \) to \( t+1 \):

\[
R(t+1) = \frac{P(t+1) - P(t)}{P(t)} = \frac{P(t+1)}{P(t)} - 1
\]

This return is then shifted by one time step:

\[
R(t+1) \longrightarrow R(t)
\]

Next, two thresholds are calculated based on the 33rd percentile (\( Q_{0.33} \)) and the 67th percentile (\( Q_{0.67} \)) of the shifted return distribution:

\[
\text{Lower Threshold} = Q_{0.33} \quad \text{and} \quad \text{Upper Threshold} = Q_{0.67}
\]

Using these thresholds, the target variable \( T(t) \) is defined as follows:

\[
T(t) = 
\begin{cases} 
-1 & \text{if} \, R(t) < Q_{0.33} \\
1 & \text{if} \, R(t) > Q_{0.67} \\
0 & \text{otherwise}
\end{cases}
\]

This classification of returns into three categories allows us to predict future market movements.

\bigskip
The graph would show the next-period returns on the y-axis and the time on the x-axis. The lower and upper thresholds \( Q_{0.33} \) and \( Q_{0.67} \) would be highlighted as horizontal lines. Points below \( Q_{0.33} \) would be marked as -1 (downward trend), points above \( Q_{0.67} \) as 1 (upward trend), and points between the two as 0 (neutral).

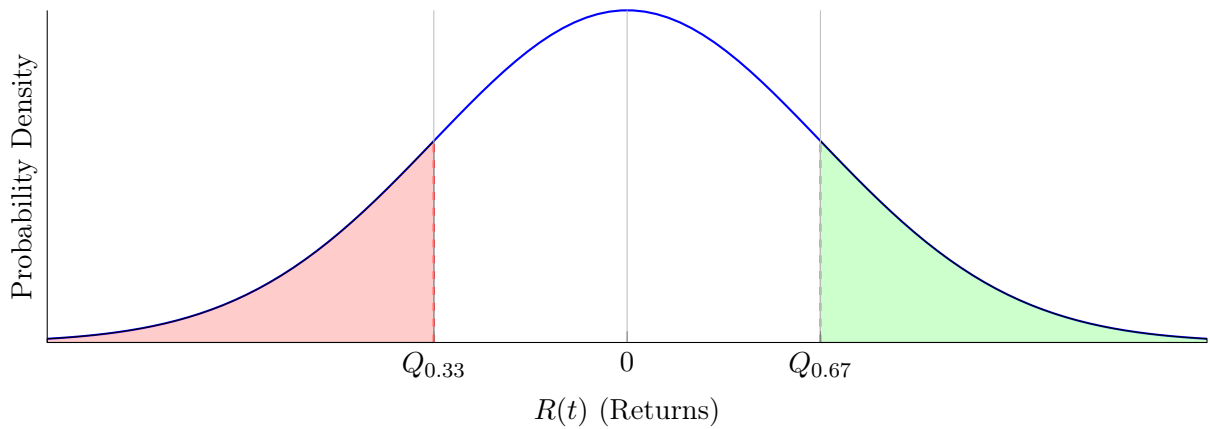
\begin{figure}[h!]
    \centering
    \begin{tikzpicture}
        \begin{axis}[
            domain=-3:3, 
            samples=100, 
            ymin=0, 
            axis lines*=left, 
            xlabel={$R(t)$ (Returns)},
            ylabel={Probability Density},
            height=6cm, 
            width=17cm,
            xtick={-1, 0, 1}, 
            xticklabels={$Q_{0.33}$, 0, $Q_{0.67}$},
            ytick=\empty,
            enlargelimits=false, 
            clip=false, 
            axis on top,
            grid = major
        ]
        
        % Plot the normal distribution
        \addplot[blue, thick] {1/sqrt(2*pi) * exp(-x^2/2)};
        
        % Add shaded area for lower threshold (Q_0.33)
        \addplot [fill=red, fill opacity=0.2, domain=-3:-1] {1/sqrt(2*pi) * exp(-x^2/2)} \closedcycle;
        
        % Add shaded area for upper threshold (Q_0.67)
        \addplot [fill=green, fill opacity=0.2, domain=1:3] {1/sqrt(2*pi) * exp(-x^2/2)} \closedcycle;
        
        % Vertical lines for the thresholds
        \draw[dashed, thick, red] (-1,0) -- (-1,{1/sqrt(2*pi) * exp(-1^2/2)});
        \draw[dashed, thick, green] (1,0) -- (1,{1/sqrt(2*pi) * exp(-1^2/2)});
        
        \end{axis}
    \end{tikzpicture}
    \caption{Normal Distribution of Returns with 33rd and 67th Percentiles}
\end{figure}

\begin{itemize}
  \item The red line indicates the lower threshold \( Q_{0.33} \).
  \item The green line indicates the upper threshold \( Q_{0.67} \).
  \item The blue curve represents the next-period returns \( R(t) \).
  \item Points above the green line are classified as \( +1 \), points below the red line as \( -1 \), and points between as \( 0 \).
\end{itemize}

Let $\mathbf{X} = [x_1, x_2, \dots, x_n]$ be the feature matrix, where each $x_i$ represents a feature vector. To standardize the features, we apply the following transformation:

\[
x_i^{\text{scaled}} = \frac{x_i - \mu_i}{\sigma_i}
\]

where $\mu_i$ and $\sigma_i$ are the mean and standard deviation of the feature $x_i$, respectively. The result is the scaled feature matrix $\mathbf{X}^{\text{scaled}}$.

\bigskip
Given the feature matrix $\mathbf{X}^{\text{scaled}}$ and the target vector $\mathbf{y}$, we use a set of models $\mathcal{M} = \{M_1, M_2, \dots, M_k\}$ to predict the target. For each model $M_j$, we perform $K$-fold time series cross-validation (TSCV) to select the best hyperparameters $\theta_j^*$:

\[
\theta_j^* = \arg \max_{\theta_j} \frac{1}{K} \sum_{k=1}^{K} \text{accuracy}(M_j(\mathbf{X}_k^{\text{train}}, \theta_j), \mathbf{y}_k^{\text{test}})
\]

where $\mathbf{X}_k^{\text{train}}$ and $\mathbf{y}_k^{\text{test}}$ are the training and testing sets for fold $k$.

For the best model $M_j^*$, we compute the predicted signals $\hat{y}$:

\[
\hat{y} = M_j^*(\mathbf{X}^{\text{scaled}})
\]

The strategy returns for the model are calculated by multiplying the predicted signals with the actual returns $r_t$ at each time step $t$:

\[
r_t^{\text{strategy}} = r_t \cdot \hat{y}_t
\]

The cumulative returns for the strategy are then given by:

\[
R_T^{\text{strategy}} = \prod_{t=1}^{T} (1 + r_t^{\text{strategy}})
\]

Similarly, a random strategy is defined by random signals $\hat{y}_t^{\text{random}}$, and its cumulative returns are:

\[
R_T^{\text{random}} = \prod_{t=1}^{T} (1 + r_t \cdot \hat{y}_t^{\text{random}})
\]

We obtain the results in [\textbf{Appendix \ref{appendix:500obs}}] and [\textbf{Appendix \ref{appendix:3000obs}}]

\bigskip
Now we'd like to see how the results evolve as the number of observations increases. We perform the above steps iteratively for different dataset sizes $n_{\text{obs}}$. For each size, we calculate the final cumulative returns $R_T^{\text{strategy}}$ for each model $M_j^*$ and visualize the results using a heatmap.

\begin{figure}[H]
    \centering
   \includegraphics[width=1\linewidth]{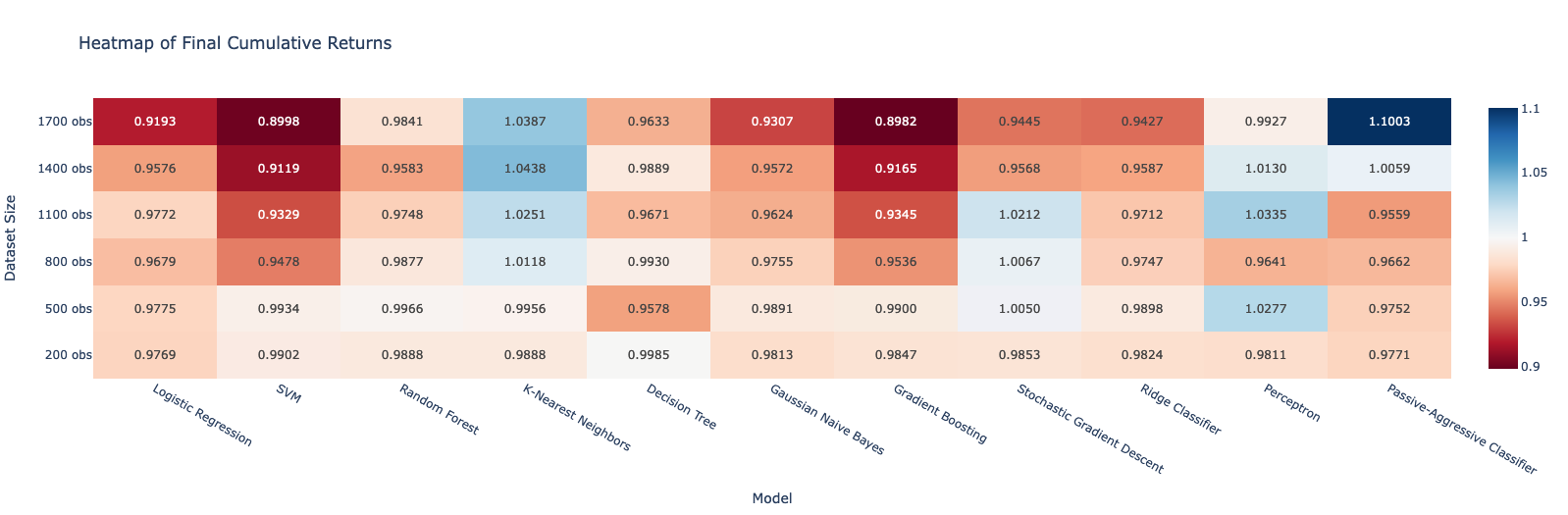}
    \caption{Evolution of model performance on increasing data}
    \label{fig:1}
\end{figure}

We observe that as the dataset size increases, the model performance deteriorates. This suggests that the signal-to-noise ratio decreases with larger datasets.This highlights that training on small data is inherently subject to the high variance bias.

The bias-variance tradeoff can be expressed as follows:

\[
\text{Error}(x) = \text{Bias}^2 + \text{Variance} + \text{Irreducible Error}
\]

When dealing with small datasets, the model tends to have high variance, as it overfits the limited data, leading to more variability in predictions. Specifically, the variance term grows due to overfitting on small data:

\[
\hat{f}(x) = \mathbb{E}[\hat{f}(x)] + \epsilon(x)
\]

Where:

\begin{itemize}
  \item \(\hat{f}(x)\) is the estimated model,
  \item \(\mathbb{E}[\hat{f}(x)]\) is the expected prediction,
  \item \(\epsilon(x)\) is the error due to overfitting.
\end{itemize}

\section{Multivariate Time Series Classification with Big Data}

\subsection{Overview of Big Data in Financial Time Series}

Big data in financial markets involves vast datasets that are generated at high speed, from a multitude of sources, often with complex structures. The definition of big data often revolves around the 3 \textbf{V}'s : \textbf{Volume}, \textbf{Velocity}, and \textbf{Variety} \cite{goldstein2021big}. Some definitions extend this to include \textbf{Veracity} and \textbf{Value}.

\paragraph{Volume} refers to the sheer amount of data being generated. Financial markets produce massive datasets, such as tick-by-tick trading data, where every single trade across global markets is recorded in real time.

\paragraph{Velocity} is the speed at which this data is produced and processed. For example, high-frequency trading generates data at millisecond intervals, which requires rapid processing and analysis.

\paragraph{Variety} addresses the different types of data available in financial markets. This can include structured data, like stock prices and volumes, and unstructured data, such as news articles, tweets, or financial reports.

\paragraph{Veracity} refers to the uncertainty and quality of data. In financial markets, the data may be incomplete, noisy, or biased, which adds an extra layer of complexity.

\paragraph{Value} highlights the potential insights that can be drawn from this data to improve financial decision-making.

\bigskip
In big data contexts, the \textbf{MapReduce} paradigm is often employed to handle large-scale data processing by dividing computations into parallel tasks. This framework, inspired by Google’s infrastructure, is particularly well-suited for distributed environments like Hadoop clusters, enabling efficient data processing across many nodes \cite{kumar2013verification}.

\subsubsection{Comparison with Small Data}

In contrast to big data, small data is much more manageable and less complex. Machine learning algorithms trained on small data typically generalize poorly on larger datasets. With small data, models such as SVMs can handle datasets in the range of thousands or hundreds of thousands of data points, whereas big data often contains millions or billions of data points.

For example, in financial markets, an SVM trained on a small dataset (e.g., daily closing prices of a single stock over 10 years) will not scale effectively to high-frequency trading data for thousands of stocks. In big data contexts, parallelization and distributed processing become critical for handling such vast datasets \cite{christ2016distributed}.

\subsubsection{Examples in Financial Markets}

One simple example of big data in financial markets is the analysis of Bitcoin (BTC) trading data. Suppose we analyze Bitcoin from 2012 to 2022 on a minute timeframe, with 200 features (such as price, volume, technical indicators, etc.). The dataset size can be approximated as:

\[
\text{Time frame} = 10 \text{ years} \times 365 \times 24 \times 60 \approx 5.26 \times 10^6 \text{ minutes.}
\]
\[
\text{Data points} = 5.26 \times 10^6 \times 200 = 1.05 \times 10^9 \text{ data points.}
\]

If we assume an SVM algorithm takes 1 second to process 1000 data points, the time to train the algorithm on this dataset would be:

\[
T_{\text{train}} = \frac{1.05 \times 10^9}{10^3} \times 1 \text{ sec} = 1.05 \times 10^6 \text{ seconds} \approx 12.15 \text{ days.}
\]

For more complex machine learning algorithms not designed for big data, the training time can become prohibitively large. A comparison of training times for various algorithms on this dataset is shown in Table \ref{tab:train_times}.

\begin{table}[h]
\centering
\begin{tabular}{|c|c|}
\hline
\textbf{Algorithm} & \textbf{Estimated Training Time} \\
\hline
SVM (linear kernel) & 12.15 days \\
Decision Tree & 4.1 days \\
Random Forest & 22.5 days \\
Deep Learning (simple) & 5.2 days \\
\hline
\end{tabular}
\caption{Estimated Training Time for Various Algorithms on a Bitcoin Dataset (Minute-level, 200 Features)}
\label{tab:train_times}
\end{table}

This illustrates the challenges of applying traditional machine learning techniques to big data and underscores the importance of scalable algorithms and infrastructure in financial time series analysis.

\subsection{Advanced Machine Learning Techniques for Big Data}

\subsubsection{Deep Learning Approaches}

We worked on \textbf{ConvTimeNet}: A Deep Hierarchical Fully Convolutional Model for Multivariate Time Series Analysis \cite{cheng2024convtimenet}. This model is specifically designed for big data scenarios in time series analysis. ConvTimeNet employs fully convolutional layers to capture both short- and long-term dependencies in time series. The key components of ConvTimeNet are the \textbf{deformable patch embedding} and \textbf{fully convolutional blocks}, allowing the model to learn multi-scale representations:

\[
\text{DePatch}(X) = \sum_{i=1}^N \text{Patch}_i(X)
\]
\[
F(x) = \text{Conv}(W \cdot x + b)
\]

The deformable patch embedding selects time points based on data-driven approaches, ensuring flexibility in feature extraction. The model's fully convolutional blocks then apply deepwise and pointwise convolutions to learn both temporal and cross-variable dependencies, making it highly effective for multivariate time series analysis.

\begin{figure}[H]
    \centering
   \includegraphics[width=0.6\linewidth]{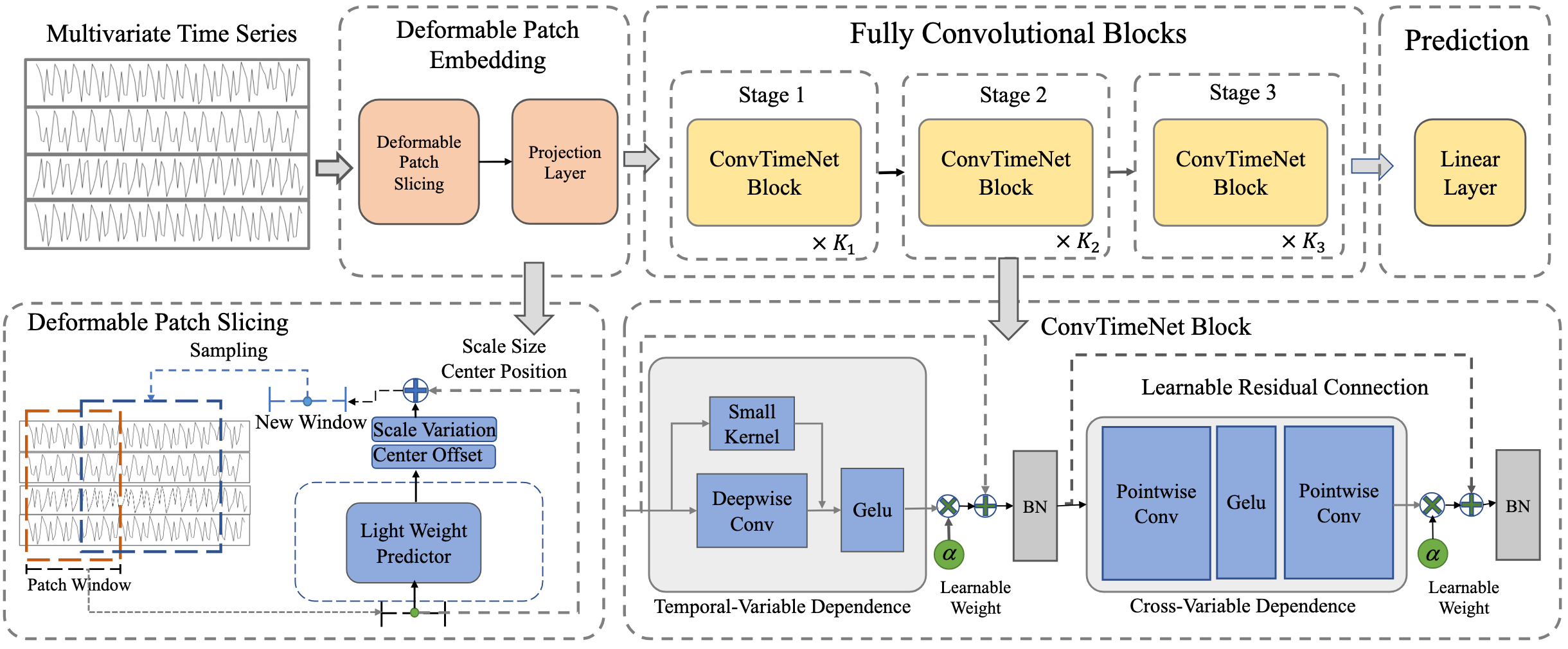}
    \caption{Illustration of the newly proposed ConvTimeNet by \cite{cheng2024convtimenet}}
    \label{fig:1}
\end{figure}

\subsubsection{Scalability of Algorithms}

In big data scenarios, the scalability of machine learning algorithms becomes crucial. Algorithms like ConvTimeNet scale efficiently with data size due to the convolutional operations' parallelizability. The time complexity is significantly reduced compared to traditional models, like LSTMs, which suffer from quadratic complexity:

\[
\mathcal{O}(n^2) \quad \text{(for self-attention)} \quad vs. \quad \mathcal{O}(n \cdot k) \quad \text{(for convolutional layers)}
\]

This reduction in complexity allows ConvTimeNet to handle vast datasets more efficiently while maintaining high predictive accuracy.

\subsubsection{Handling Multivariate Data}

ConvTimeNet excels at handling multivariate data by modeling both temporal and cross-variable dependencies. The combination of deepwise and pointwise convolutional layers ensures that the model captures the intricate relationships between multiple variables. Given a multivariate input \(X \in \mathbb{R}^{C \times T}\) (with \(C\) variables and \(T\) time points), ConvTimeNet applies convolutions over both dimensions:

\[
h_t = \sum_{c=1}^C W_c * x_{t,c}
\]

This allows the model to learn both local and global patterns in the data, making it well-suited for high-dimensional financial datasets.

\subsection{Application on EUR/USD}

\subsubsection{Theory}

We apply a deep learning model, the ConvTimeNet architecture \cite{cheng2024convtimenet}, to forecast EUR/USD exchange rates, whose code is in [\textbf{Appendix \ref{appendix:convTimeNet}}]. The key advantage of this model is its ability to capture both local and global patterns in time series using convolutional layers.

\begin{figure}[h]
    \centering
    \includegraphics[width=0.9\linewidth]{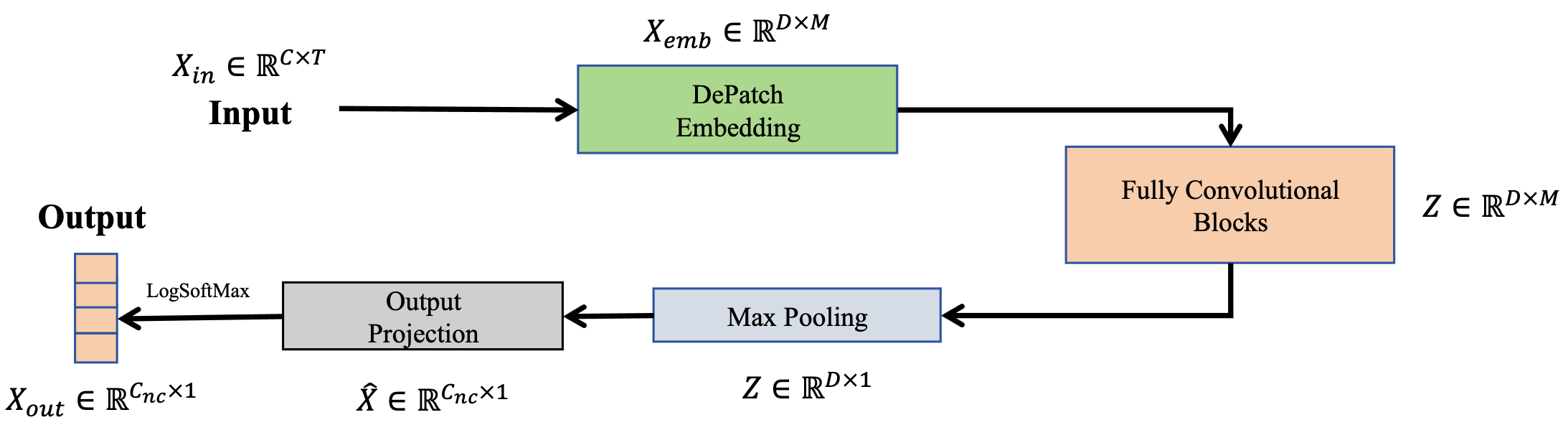}
    \caption{ConvTimeNet Architecture}
    \label{fig:conv_architecture}
\end{figure}

The architecture involves the following main components:
\begin{itemize}
    \item \textbf{Deformable Patch Embedding}: Converts the input data \( X_{\text{in}} \in \mathbb{R}^{C \times T} \) into a representation \( X_{\text{emb}} \in \mathbb{R}^{D \times M} \), where \( C \) is the number of features and \( T \) is the length of the time series.
    \item \textbf{Fully Convolutional Blocks}: These blocks consist of convolutional layers that apply filters to capture temporal dependencies:
    \[
    F(x) = \text{Conv}(W \cdot x + b)
    \]
    where \( W \) is the filter matrix and \( b \) is the bias.
    \item \textbf{Output Projection}: The final output is obtained by applying a \textit{log-softmax} function to produce the predicted class probabilities:
    \[
    \hat{y} = \log \frac{\exp(y_i)}{\sum_{j=1}^{nc} \exp(y_j)}
    \]
    where \( nc \) represents the number of classes.
\end{itemize}

The loss function used for optimization is the cross-entropy loss:
\[
\mathcal{L}(\hat{y}, y) = - \sum_{i=1}^{nc} y_i \log(\hat{y}_i)
\]
where \( y_i \) is the true class label and \( \hat{y}_i \) is the predicted probability for class \( i \).

\subsubsection{Results}

\begin{figure}[h]
    \centering
    \includegraphics[width=1\linewidth]{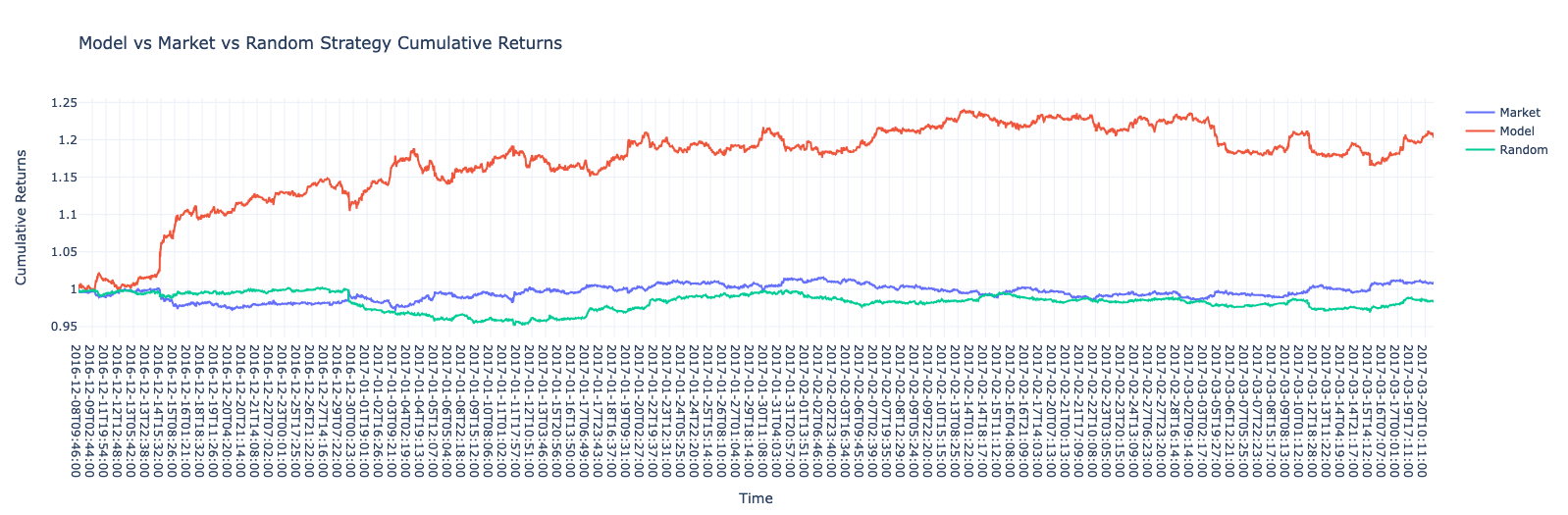}
    \caption{Model vs Market vs Random Strategy Cumulative Returns}
    \label{fig:cumulative_returns}
\end{figure}

We applied this model to EUR/USD data using 100,000 minute-level samples with 200 features. The training was carried out using the \textit{Adam} optimizer and cross-entropy loss over 100 epochs. Figure \ref{fig:conv_architecture} shows the architecture of ConvTimeNet, and Figure \ref{fig:cumulative_returns} displays the cumulative returns comparison between the model, the market, and a random strategy. 

The performance of the model was evaluated by simulating a trading strategy based on the predicted signals. The returns from the model's predictions are defined as:

\[
r_t^{\text{strategy}} = r_t \cdot \hat{y}_t
\]

where \(r_t\) is the actual return, and \(\hat{y}_t\) is the predicted signal.

As shown in the results, the model outperformed the random strategy, generating a final return of 20\% while the market remained relatively flat. This indicates that the model was able to identify profitable patterns in the EUR/USD data, as demonstrated by the cumulative returns curve.

\cleardoublepage

\section{Conclusion}

We stated earlier that waiting longer in the market yields more data, which in turn can improve the accuracy of trading action classification, potentially allowing us to beat the market. However, as highlighted by Andrew Lo in the Adaptive Markets Hypothesis \cite{lo2004adaptive}, the longer we wait, the more the market adapts to these predictions, making it harder to outsmart. This poses a critical question: Is more data truly beneficial, or does the market's evolution limit the effectiveness of data-driven predictions? 

One possibility is that both are correct in different contexts. As data accumulates, especially with the rise of big data, models tend to overfit in small data environments but may perform better in big data settings. Nevertheless, the adaptive nature of the markets complicates this.

\subsection{Non-Euclidean Spaces and Neural Networks}

Another interesting avenue for future research is the application of deep learning in non-Euclidean spaces. Most deep learning models assume that data lies in a Euclidean space, but recent work on Graph Neural Networks (GNNs) \cite{asif2021graph} highlights the importance of non-Euclidean structures in various domains, including finance. In financial markets, the relationships between assets could be better captured using graphs, and applying deep learning to these graphs could provide more accurate predictions. A typical Euclidean-based model learns weights as:
\[
h_t = \sigma(W_h h_{t-1} + W_x x_t + b_h)
\]
However, in non-Euclidean spaces, the convolutional operator generalizes to graph structures, where the neighborhood function $N(v)$ influences updates.

\subsection{Parallelization with Spark and GPU Acceleration}

When dealing with big data, traditional CPU-based computations become infeasible. Apache Spark has emerged as a powerful framework for distributed computing, enabling machine learning models to scale across large datasets with parallel processing \cite{lunga2020apache}. Furthermore, GPU acceleration drastically reduces training time, especially for complex models like CNNs and LSTMs. A standard GPU-accelerated training loop, for instance, allows matrix computations to be handled in parallel:
\[
\mathcal{O}(n^2) \quad \text{(for dense layers)} \quad vs. \quad \mathcal{O}(n \cdot k) \quad \text{(for convolutional layers)}
\]
This provides the scalability necessary for handling minute-level datasets over years of market data.

\bigskip
Thus, as we advance in financial time series classification, the balance between data accumulation, market adaptation, and technological advancements such as non-Euclidean networks and parallel computing must be carefully considered. The answer may lie not in more data alone, but in smarter algorithms and better utilization of available computational resources.

\cleardoublepage

\appendix

\section{\texttt{strikeXGBoost}}
\label{appendix:strikeXGBoost}
\begin{lstlisting}[language=Python]
import requests
import pandas as pd
import numpy as np
import xgboost as xgb
from sklearn.preprocessing import StandardScaler
from sklearn.model_selection import TimeSeriesSplit, GridSearchCV
from sklearn.metrics import accuracy_score
import plotly.graph_objs as go
import warnings
warnings.simplefilter(action='ignore')

url = "https://api.tokeninsight.com/api/v1/history/coins/strike-strike?length=-1"

headers = {
    "accept": "application/json",
    "TI_API_KEY": ""
}
response = requests.get(url, headers=headers)
data = response.json()
market_chart = data['data']['market_chart']
df = pd.DataFrame(market_chart, columns=['timestamp', 'price'])
df['timestamp'] = pd.to_datetime(df['timestamp'], unit='ms')
df.set_index('timestamp', inplace=True)

df['returns'] = df['price'].pct_change().fillna(0)
df['SMA_10'] = df['price'].rolling(window=10).mean().fillna(df['price'].mean())
df['SMA_30'] = df['price'].rolling(window=30).mean().fillna(df['price'].mean())

df['target'] = np.where(df['returns'].shift(-1) > 0, 1, 0)
features = df[['returns', 'SMA_10', 'SMA_30']].iloc[:-1] 
target = df['target'].iloc[:-1]

scaler = StandardScaler()
features_scaled = scaler.fit_transform(features)

tscv = TimeSeriesSplit(n_splits=5)

xgb_model = xgb.XGBClassifier(use_label_encoder=False, eval_metric='mlogloss')
params = {
    'n_estimators': [50, 100],
    'learning_rate': [0.01, 0.1],
    'max_depth': [3, 5]
}

best_model = None
best_accuracy = 0
for train_index, test_index in tscv.split(features_scaled):
    X_train, X_test = features_scaled[train_index], features_scaled[test_index]
    y_train, y_test = target.iloc[train_index], target.iloc[test_index]
    
    grid_search = GridSearchCV(xgb_model, params, cv=3, scoring='accuracy')
    grid_search.fit(X_train, y_train)
    y_pred = grid_search.best_estimator_.predict(X_test)
    
    accuracy = accuracy_score(y_test, y_pred)
    if accuracy > best_accuracy:
        best_model = grid_search.best_estimator_
        best_accuracy = accuracy

predicted_signals = best_model.predict(features_scaled)

df['strategy_returns'] = df['returns'].iloc[:-1] * predicted_signals

random_strategy = np.random.choice([0, 1], size=len(df)-1)
df['random_strategy_returns'] = df['returns'].iloc[:-1] * random_strategy

cumulative_returns = {
    'Market': (1 + df['returns']).cumprod(),
    'Model': (1 + df['strategy_returns']).cumprod(),
    'Random': (1 + df['random_strategy_returns']).cumprod()
}

fig = go.Figure()
for key, values in cumulative_returns.items():
    fig.add_trace(go.Scatter(x=df.index, y=values, mode='lines', name=f'{key}'))

fig.update_layout(
    title='XGBoost Model Strategy Backtest',
    xaxis_title='Time',
    yaxis_title='Cumulative Returns',
    template='plotly_white'
)
fig.show()


\end{lstlisting}

\section{\texttt{smallDataApplication}}
\label{appendix:smallDataApplication}
\begin{lstlisting}[language=Python]

import pandas as pd

df = pd.read_csv("data.csv")
df = df.drop(['next_return', 'return_t-0'], axis=1)
df['returns'] = df['close'].pct_change()
df['returns'].fillna(0, inplace=True)
df = df.set_index('date')
df_min = df.iloc[:2000]


import pandas as pd
from sklearn.preprocessing import StandardScaler
from sklearn.model_selection import TimeSeriesSplit
import warnings
from sklearn.exceptions import ConvergenceWarning
import numpy as np
import pandas as pd
from sklearn.model_selection import GridSearchCV
from sklearn.metrics import confusion_matrix, classification_report, accuracy_score
import plotly.graph_objs as go
from plotly.subplots import make_subplots

warnings.simplefilter(action='ignore', category=ConvergenceWarning)
warnings.simplefilter(action='ignore', category=pd.errors.PerformanceWarning)
warnings.simplefilter(action='ignore', category=pd.errors.SettingWithCopyWarning)

features = df_min.drop(columns=['target'])
target = df_min['target']

scaler = StandardScaler()
features_scaled = scaler.fit_transform(features)

tscv = TimeSeriesSplit(n_splits=5)

split_index = int(len(features_scaled) * 0.8)
X_train, X_test = features_scaled[:split_index], features_scaled[split_index:]
y_train, y_test = target[:split_index], target[split_index:]


import numpy as np
import pandas as pd
from sklearn.model_selection import TimeSeriesSplit, GridSearchCV
from sklearn.metrics import confusion_matrix, classification_report, accuracy_score
from sklearn.preprocessing import StandardScaler
from sklearn.linear_model import LogisticRegression
import plotly.graph_objs as go
from sklearn.svm import SVC
from sklearn.ensemble import RandomForestClassifier, GradientBoostingClassifier
from sklearn.neighbors import KNeighborsClassifier
from sklearn.tree import DecisionTreeClassifier
from sklearn.naive_bayes import GaussianNB
from sklearn.linear_model import SGDClassifier, RidgeClassifier, Perceptron, PassiveAggressiveClassifier




import pandas as pd
from sklearn.preprocessing import StandardScaler
from sklearn.model_selection import TimeSeriesSplit
import warnings
from sklearn.exceptions import ConvergenceWarning
import numpy as np
import pandas as pd
from sklearn.model_selection import GridSearchCV
from sklearn.metrics import confusion_matrix, classification_report, accuracy_score
import plotly.graph_objs as go
from plotly.subplots import make_subplots

warnings.simplefilter(action='ignore', category=ConvergenceWarning)
warnings.simplefilter(action='ignore', category=pd.errors.PerformanceWarning)
warnings.simplefilter(action='ignore', category=pd.errors.SettingWithCopyWarning)


models = {
    'Logistic Regression': (LogisticRegression(), {'C': [0.01, 0.1, 1, 10, 100], 'solver': ['liblinear', 'saga']}),
    'SVM': (SVC(), {'C': [0.1, 1, 10], 'kernel': ['rbf', 'linear']}),
    'Random Forest': (RandomForestClassifier(), {'n_estimators': [50, 100, 200], 'max_depth': [None, 5, 10]}),
    'K-Nearest Neighbors': (KNeighborsClassifier(), {'n_neighbors': [3, 5, 7], 'weights': ['uniform', 'distance']}),
    'Decision Tree': (DecisionTreeClassifier(), {'max_depth': [None, 5, 10], 'min_samples_split': [2, 5, 10]}),
    'Gaussian Naive Bayes': (GaussianNB(), {}),
    'Gradient Boosting': (GradientBoostingClassifier(), {'n_estimators': [50, 100, 200], 'learning_rate': [0.01, 0.1, 1]}),
    'Stochastic Gradient Descent': (SGDClassifier(), {'alpha': [0.0001, 0.001, 0.01], 'penalty': ['l1', 'l2', 'elasticnet']}),
    'Ridge Classifier': (RidgeClassifier(), {'alpha': [0.1, 1, 10]}),
    'Perceptron': (Perceptron(), {'alpha': [0.0001, 0.001, 0.01], 'penalty': ['l1', 'l2', 'elasticnet']}),
    'Passive-Aggressive Classifier': (PassiveAggressiveClassifier(), {'C': [0.01, 0.1, 1, 10, 100]}),
}


df = pd.read_csv("data.csv")
df = df.drop(['next_return', 'return_t-0'], axis=1)
df['returns'] = df['close'].pct_change().fillna(0)
df = df.set_index('date')
df_min = df.iloc[:500]

features = df_min.drop(columns=['target'])
target = df_min['target']

scaler = StandardScaler()
features_scaled = scaler.fit_transform(features)

tscv = TimeSeriesSplit(n_splits=5)

for name, (model, params) in models.items():
    grid_search = GridSearchCV(model, params, cv=tscv, scoring='accuracy')
    grid_search.fit(features_scaled, target)
    best_model = grid_search.best_estimator_
    predicted_signals = best_model.predict(features_scaled)

    df_min['strategy_returns'] = df_min['returns'] * predicted_signals

    random_strategy = np.random.choice([-1, 0, 1], size=len(df_min))
    df_min['random_strategy_returns'] = df_min['returns'] * random_strategy

    cumulative_returns = {
        'Market': (1 + df_min['returns']).cumprod(),
        'Model': (1 + df_min['strategy_returns']).cumprod(),
        'Random': (1 + df_min['random_strategy_returns']).cumprod()
    }

    fig = go.Figure()
    for key, values in cumulative_returns.items():
        fig.add_trace(go.Scatter(x=df_min.index, y=values, mode='lines', name=f'{key}'))
    fig.update_layout(title=f'{name}\'s best model', xaxis_title='Time', yaxis_title='Cumulative Returns', template='plotly_white')
    fig.show()




import numpy as np
import pandas as pd
from sklearn.model_selection import TimeSeriesSplit, GridSearchCV
from sklearn.metrics import confusion_matrix, classification_report, accuracy_score
from sklearn.preprocessing import StandardScaler
from sklearn.linear_model import LogisticRegression
import plotly.graph_objs as go
from sklearn.svm import SVC
from sklearn.ensemble import RandomForestClassifier, GradientBoostingClassifier
from sklearn.neighbors import KNeighborsClassifier
from sklearn.tree import DecisionTreeClassifier
from sklearn.naive_bayes import GaussianNB
from sklearn.linear_model import SGDClassifier, RidgeClassifier, Perceptron, PassiveAggressiveClassifier




import pandas as pd
from sklearn.preprocessing import StandardScaler
from sklearn.model_selection import TimeSeriesSplit
import warnings
from sklearn.exceptions import ConvergenceWarning
import numpy as np
import pandas as pd
from sklearn.model_selection import GridSearchCV
from sklearn.metrics import confusion_matrix, classification_report, accuracy_score
import plotly.graph_objs as go
from plotly.subplots import make_subplots

warnings.simplefilter(action='ignore', category=ConvergenceWarning)
warnings.simplefilter(action='ignore', category=pd.errors.PerformanceWarning)
warnings.simplefilter(action='ignore', category=pd.errors.SettingWithCopyWarning)



models = {
    'Logistic Regression': (LogisticRegression(), {'C': [0.01, 0.1, 1, 10, 100], 'solver': ['liblinear', 'saga']}),
    'SVM': (SVC(), {'C': [0.1, 1, 10], 'kernel': ['rbf', 'linear'], 'gamma': ['scale', 'auto'], 'degree': [2, 3, 4]}),
    'Random Forest': (RandomForestClassifier(), {'n_estimators': [50, 100, 200], 'max_depth': [None, 5, 10]}),
    'K-Nearest Neighbors': (KNeighborsClassifier(), {'n_neighbors': [3, 5, 7], 'weights': ['uniform', 'distance']}),
    'Decision Tree': (DecisionTreeClassifier(), {'max_depth': [None, 5, 10], 'min_samples_split': [2, 5, 10]}),
    'Gaussian Naive Bayes': (GaussianNB(), {}),
    'Gradient Boosting': (GradientBoostingClassifier(), {'n_estimators': [50, 100, 200], 'learning_rate': [0.01, 0.1, 1]}),
    'Stochastic Gradient Descent': (SGDClassifier(), {'alpha': [0.0001, 0.001, 0.01], 'penalty': ['l1', 'l2', 'elasticnet']}),
    'Ridge Classifier': (RidgeClassifier(), {'alpha': [0.1, 1, 10]}),
    'Perceptron': (Perceptron(), {'alpha': [0.0001, 0.001, 0.01], 'penalty': ['l1', 'l2', 'elasticnet']}),
    'Passive-Aggressive Classifier': (PassiveAggressiveClassifier(), {'C': [0.01, 0.1, 1, 10, 100]}),
}


df = pd.read_csv("data.csv")
df = df.drop(['next_return', 'return_t-0'], axis=1)
df['returns'] = df['close'].pct_change().fillna(0)
df = df.set_index('date')
df_min = df.iloc[10000:12000]

features = df_min.drop(columns=['target'])
target = df_min['target']

scaler = StandardScaler()
features_scaled = scaler.fit_transform(features)

tscv = TimeSeriesSplit(n_splits=5)

for name, (model, params) in models.items():
    grid_search = GridSearchCV(model, params, cv=tscv, scoring='accuracy')
    grid_search.fit(features_scaled, target)
    best_model = grid_search.best_estimator_
    predicted_signals = best_model.predict(features_scaled)

    df_min['strategy_returns'] = df_min['returns'] * predicted_signals

    random_strategy = np.random.choice([-1, 0, 1], size=len(df_min))
    df_min['random_strategy_returns'] = df_min['returns'] * random_strategy

    cumulative_returns = {
        'Market': (1 + df_min['returns']).cumprod(),
        'Model': (1 + df_min['strategy_returns']).cumprod(),
        'Random': (1 + df_min['random_strategy_returns']).cumprod()
    }

    fig = go.Figure()
    for key, values in cumulative_returns.items():
        fig.add_trace(go.Scatter(x=df_min.index, y=values, mode='lines', name=f'{key}'))
    fig.update_layout(title=f'{name}\'s best model', xaxis_title='Time', yaxis_title='Cumulative Returns', template='plotly_white')
    fig.show()


import numpy as np
import pandas as pd
from sklearn.model_selection import TimeSeriesSplit, GridSearchCV
from sklearn.preprocessing import StandardScaler
from sklearn.linear_model import LogisticRegression
from sklearn.svm import SVC

models = {
    'Logistic Regression': (LogisticRegression(), {'C': [0.01, 0.1, 1, 10, 100], 'solver': ['liblinear', 'saga']}),
    'SVM': (SVC(), {'C': [0.1, 1, 10], 'kernel': ['rbf', 'linear'], 'gamma': ['scale', 'auto'], 'degree': [2, 3, 4]}),
    'Random Forest': (RandomForestClassifier(), {'n_estimators': [50, 100, 200], 'max_depth': [None, 5, 10]}),
    'K-Nearest Neighbors': (KNeighborsClassifier(), {'n_neighbors': [3, 5, 7], 'weights': ['uniform', 'distance']}),
    'Decision Tree': (DecisionTreeClassifier(), {'max_depth': [None, 5, 10], 'min_samples_split': [2, 5, 10]}),
    'Gaussian Naive Bayes': (GaussianNB(), {}),
    'Gradient Boosting': (GradientBoostingClassifier(), {'n_estimators': [50, 100, 200], 'learning_rate': [0.01, 0.1, 1]}),
    'Stochastic Gradient Descent': (SGDClassifier(), {'alpha': [0.0001, 0.001, 0.01], 'penalty': ['l1', 'l2', 'elasticnet']}),
    'Ridge Classifier': (RidgeClassifier(), {'alpha': [0.1, 1, 10]}),
    'Perceptron': (Perceptron(), {'alpha': [0.0001, 0.001, 0.01], 'penalty': ['l1', 'l2', 'elasticnet']}),
    'Passive-Aggressive Classifier': (PassiveAggressiveClassifier(), {'C': [0.01, 0.1, 1, 10, 100]}),
}

cum_final = {}

initial_size = 200
increment_size = 300

tscv = TimeSeriesSplit(n_splits=5)

for i in range(6):
    size = initial_size + (increment_size * i)
    df_min = df.iloc[10000:10000 + size]

    features = df_min.drop(columns=['target'])
    target = df_min['target']

    scaler = StandardScaler()
    features_scaled = scaler.fit_transform(features)

    cum_ret_dict = {}

    for name, (model, params) in models.items():
        grid_search = GridSearchCV(model, params, cv=tscv, scoring='accuracy')
        grid_search.fit(features_scaled, target)
        best_model = grid_search.best_estimator_
        predicted_signals = best_model.predict(features_scaled)

        df_min['strategy_returns'] = df_min['returns'] * predicted_signals
        model_last_cum_ret = (1 + df_min['strategy_returns']).cumprod().iloc[-1]
        cum_ret_dict[name] = model_last_cum_ret

    cum_final[f'{size} obs'] = cum_ret_dict

cum_final_df = pd.DataFrame(cum_final).T

fig = go.Figure(data=go.Heatmap(

    z=cum_final_df.values,

    x=cum_final_df.columns,

    y=cum_final_df.index,

    colorscale='RdBu',

    text=cum_final_df.applymap(lambda x: f"{x:.4f}").values,

    texttemplate="%{text}"

))



fig.update_layout(

    title="Heatmap of Final Cumulative Returns",

    xaxis_title="Model",

    yaxis_title="Dataset Size",

    template="plotly_white"

)



fig.show()


\end{lstlisting}

\section{500 Observations Results}
\label{appendix:500obs}
\begin{figure}[H]
    \centering
    \includegraphics[width=1\linewidth]{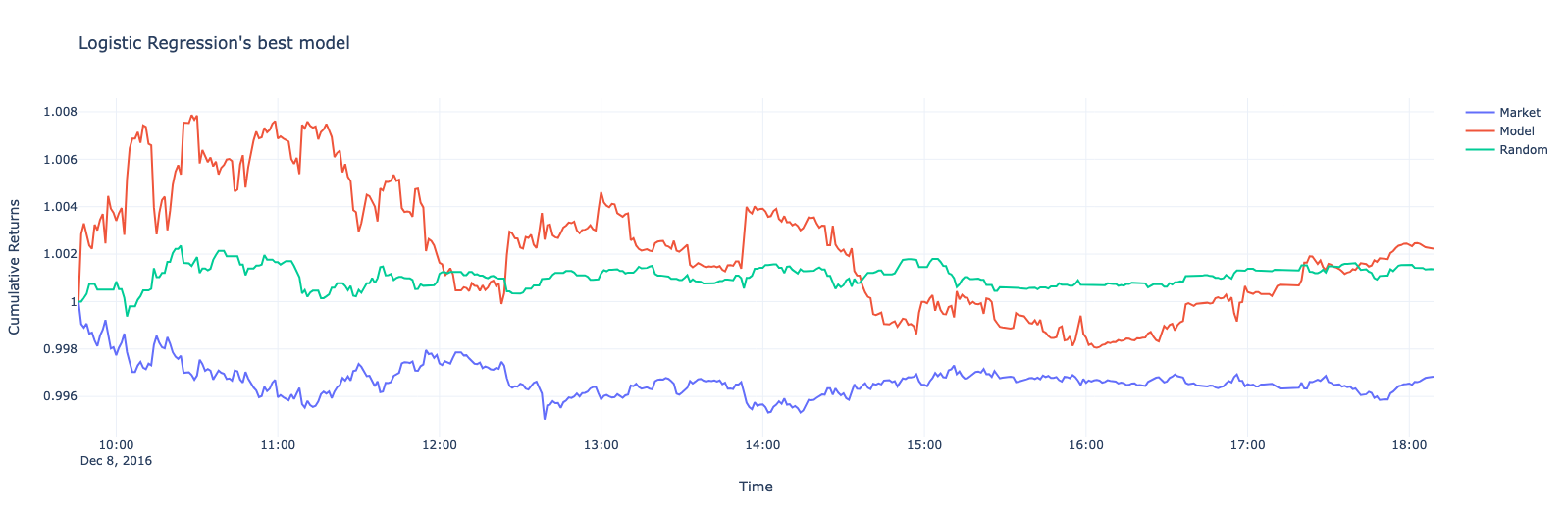}
\end{figure}
\begin{figure}[H]
    \centering
    \includegraphics[width=1\linewidth]{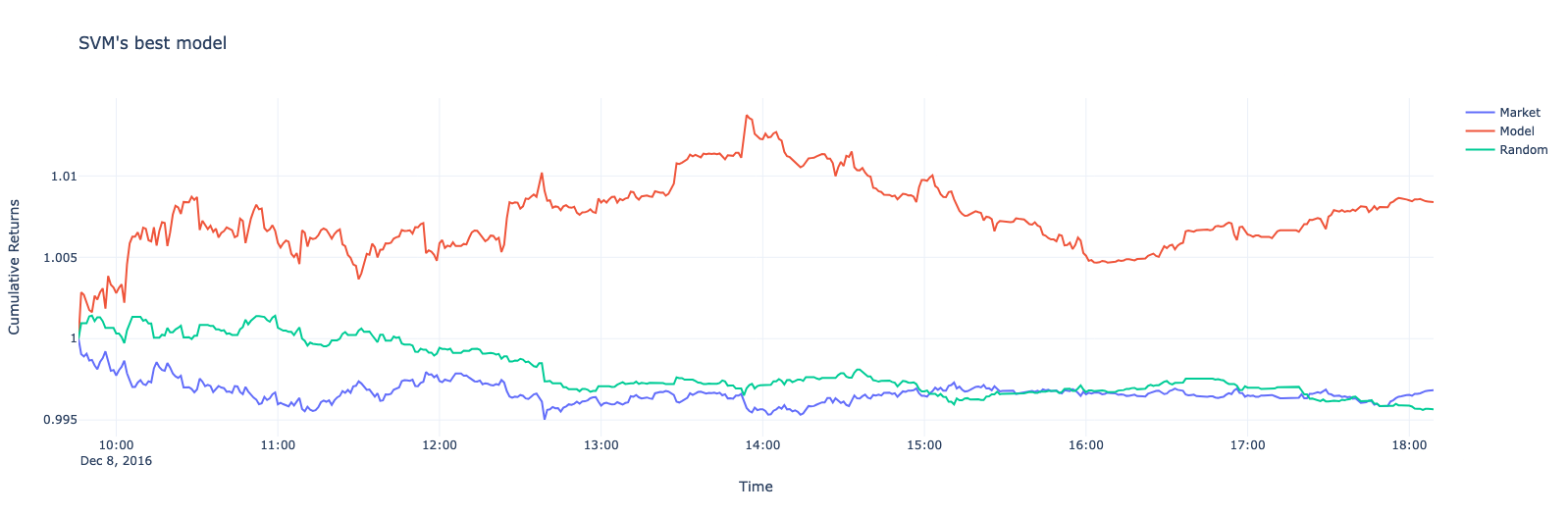}
\end{figure}
\begin{figure}[H]
    \centering
    \includegraphics[width=1\linewidth]{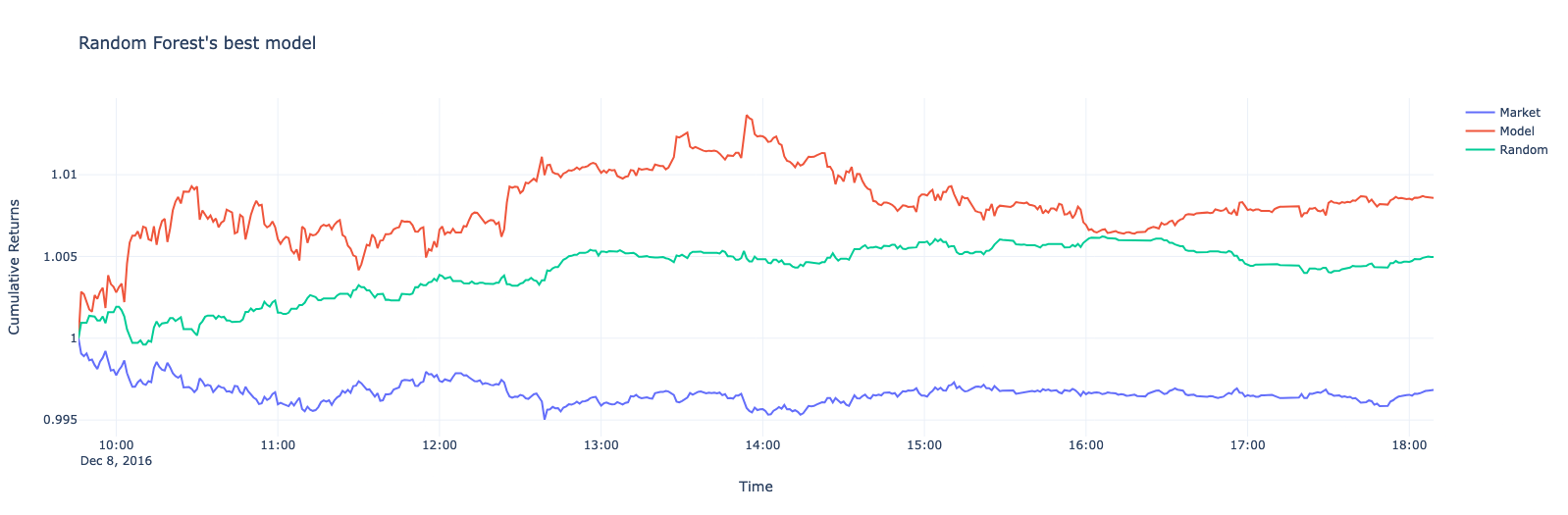}
\end{figure}
\begin{figure}[H]
    \centering
    \includegraphics[width=1\linewidth]{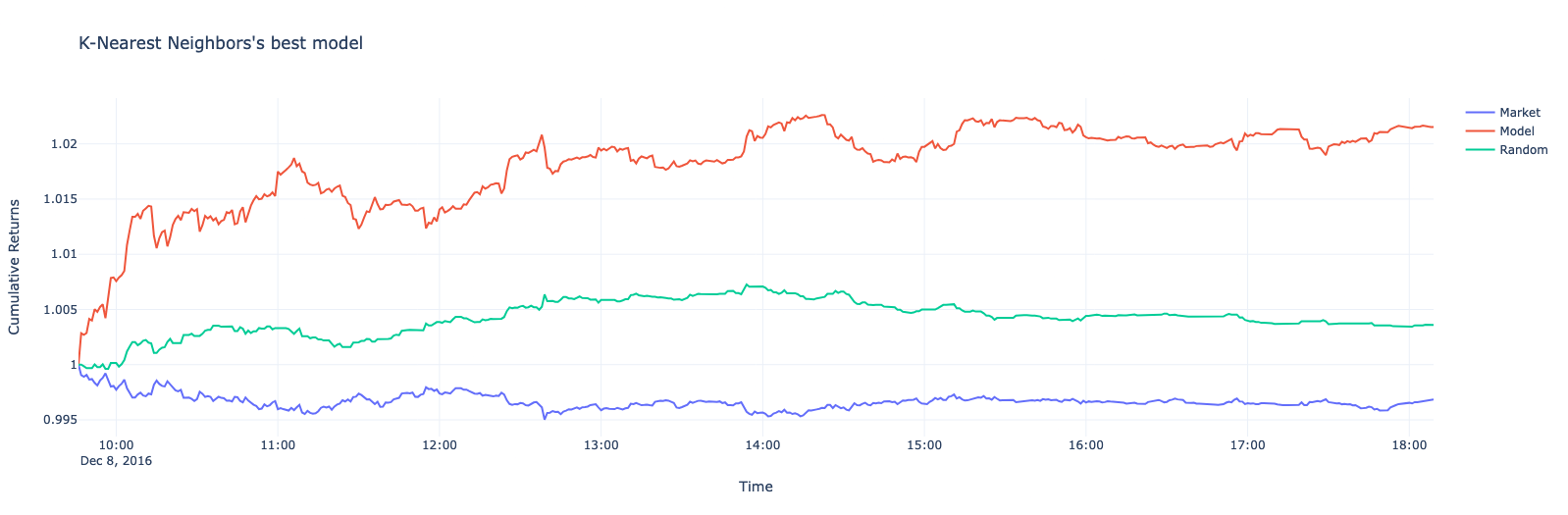}
\end{figure}
\begin{figure}[H]
    \centering
    \includegraphics[width=1\linewidth]{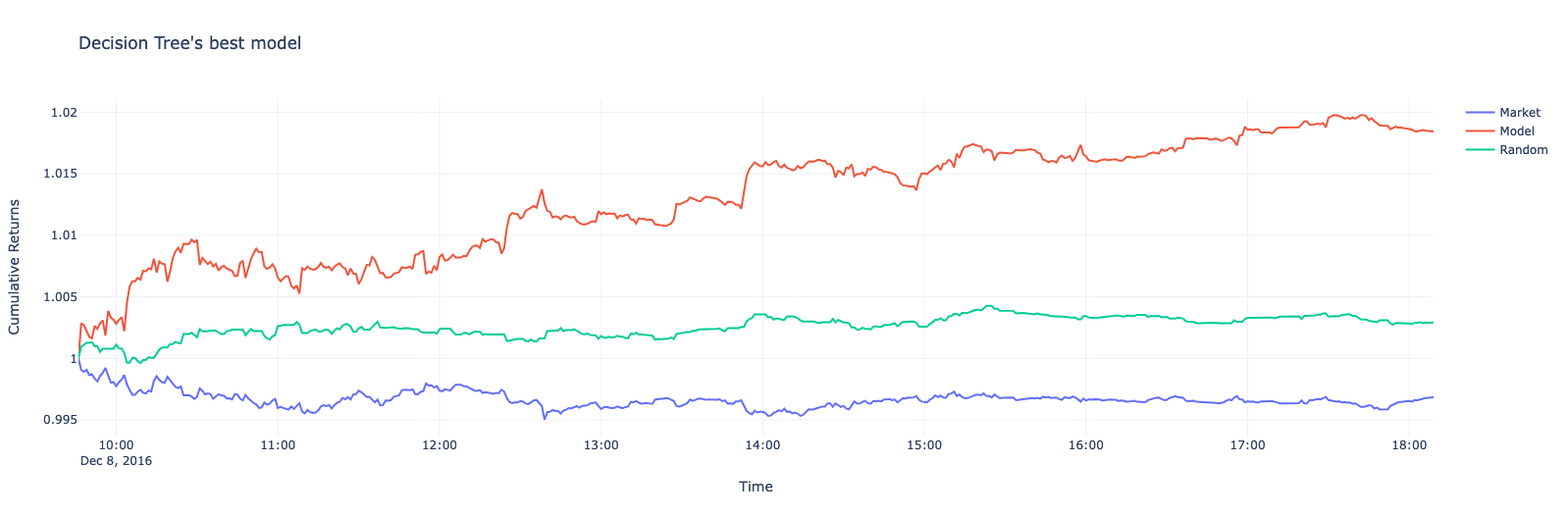}
\end{figure}
\begin{figure}[H]
    \centering
    \includegraphics[width=1\linewidth]{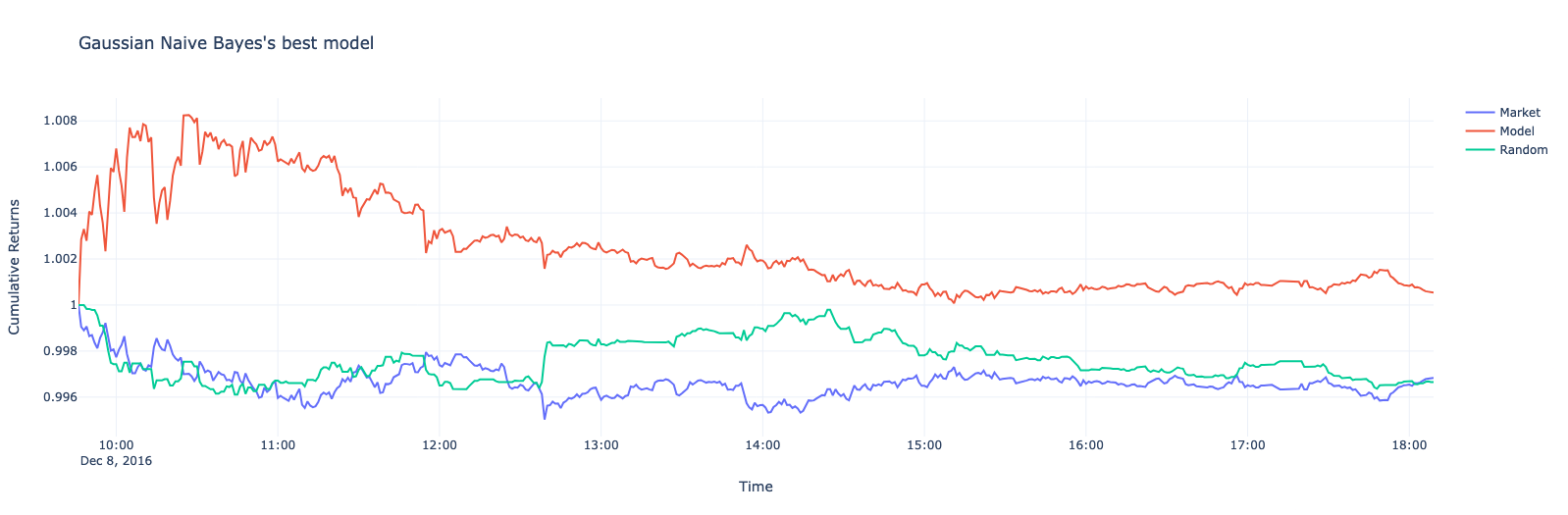}
\end{figure}
\begin{figure}[H]
    \centering
    \includegraphics[width=1\linewidth]{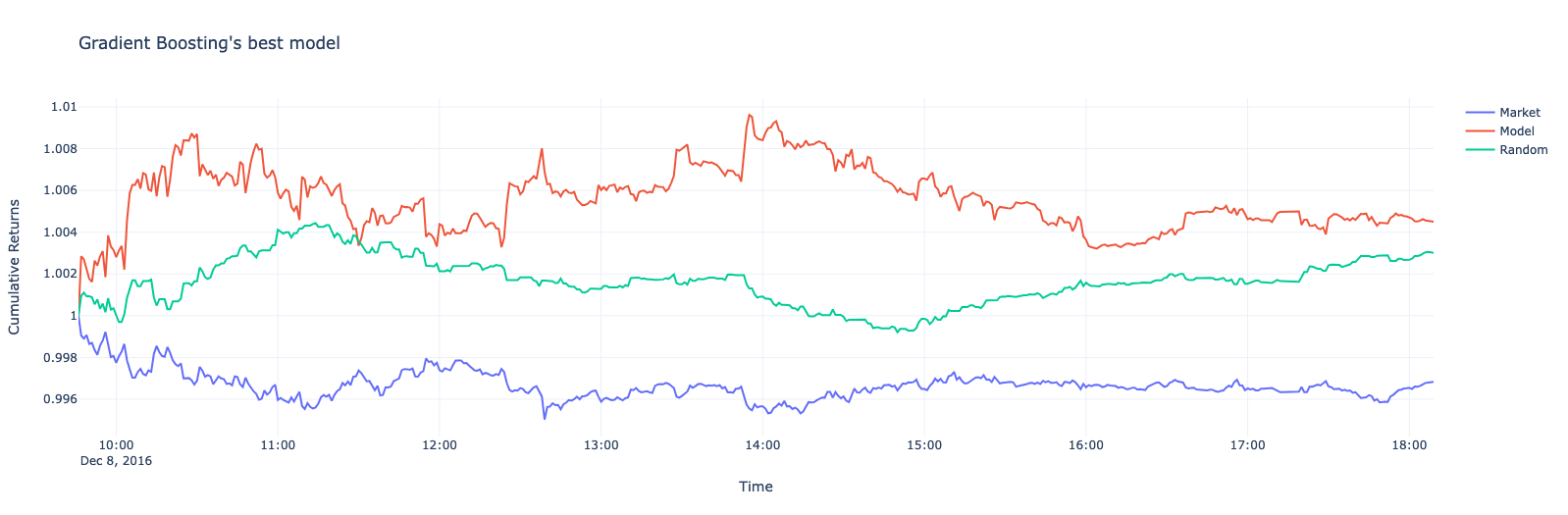}
\end{figure}
\begin{figure}[H]
    \centering
    \includegraphics[width=1\linewidth]{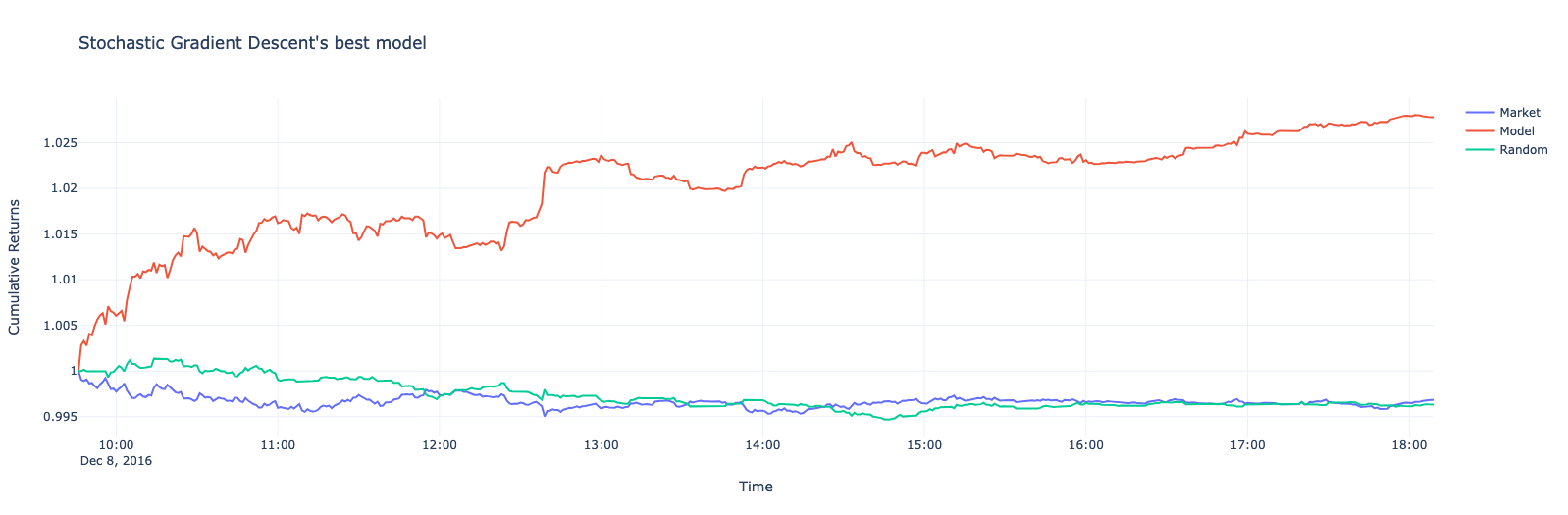}
\end{figure}
\begin{figure}[H]
    \centering
    \includegraphics[width=1\linewidth]{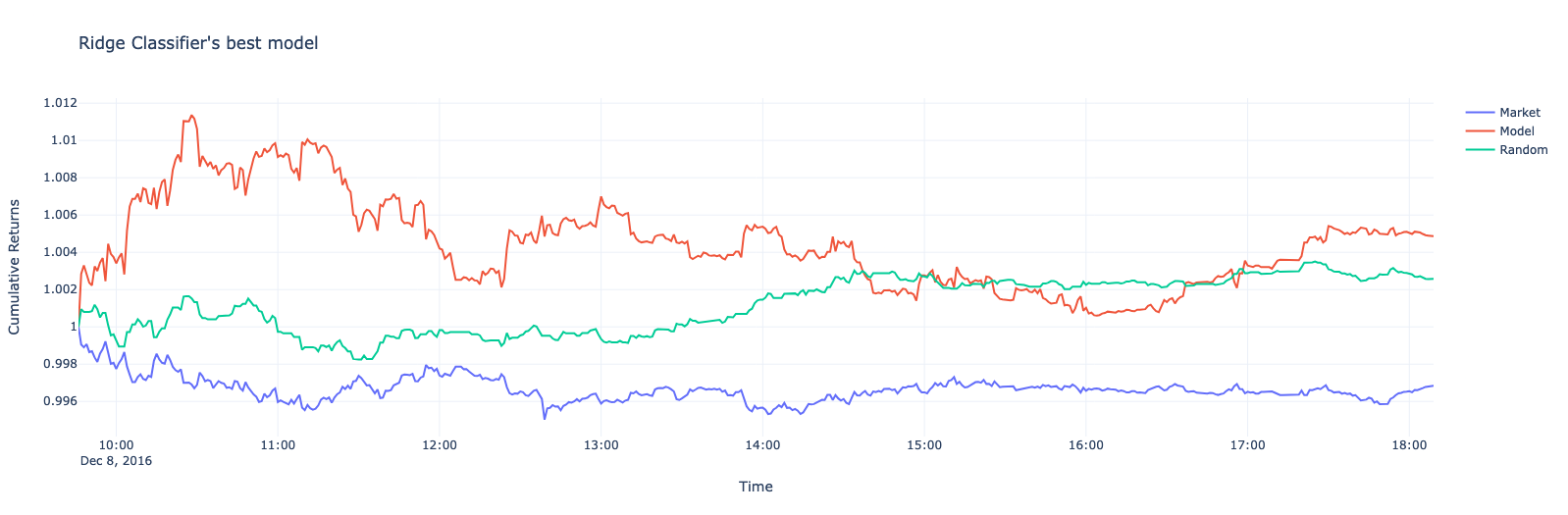}
\end{figure}
\begin{figure}[H]
    \centering
    \includegraphics[width=1\linewidth]{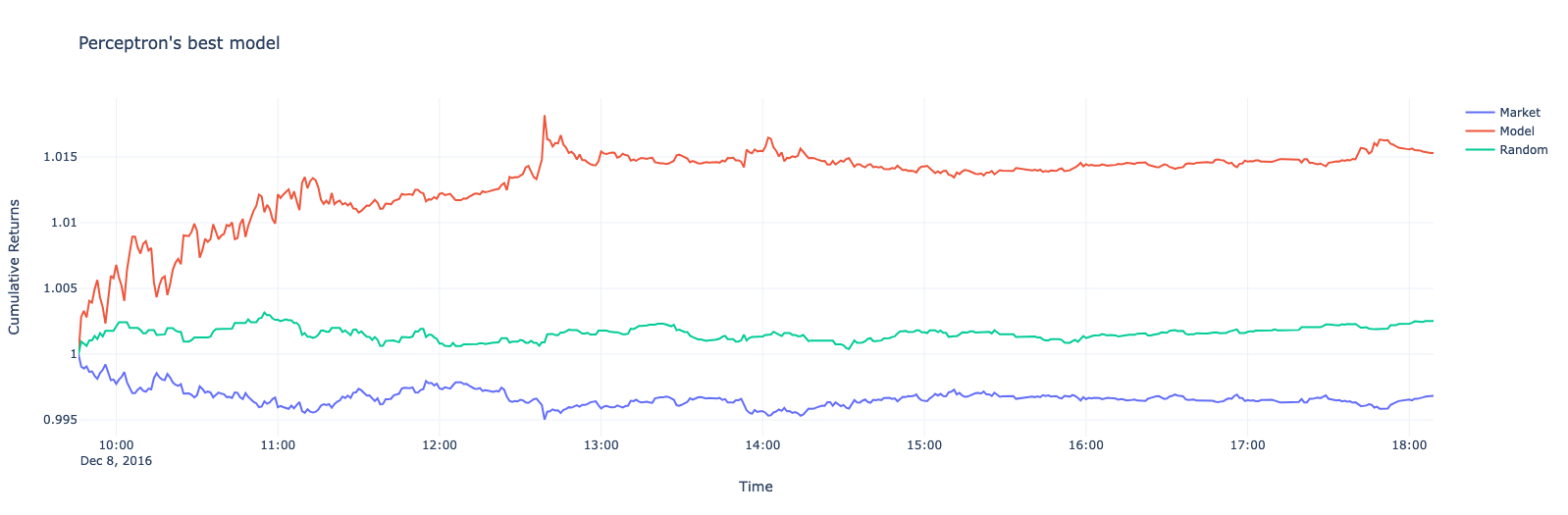}
\end{figure}
\begin{figure}[H]
    \centering
    \includegraphics[width=1\linewidth]{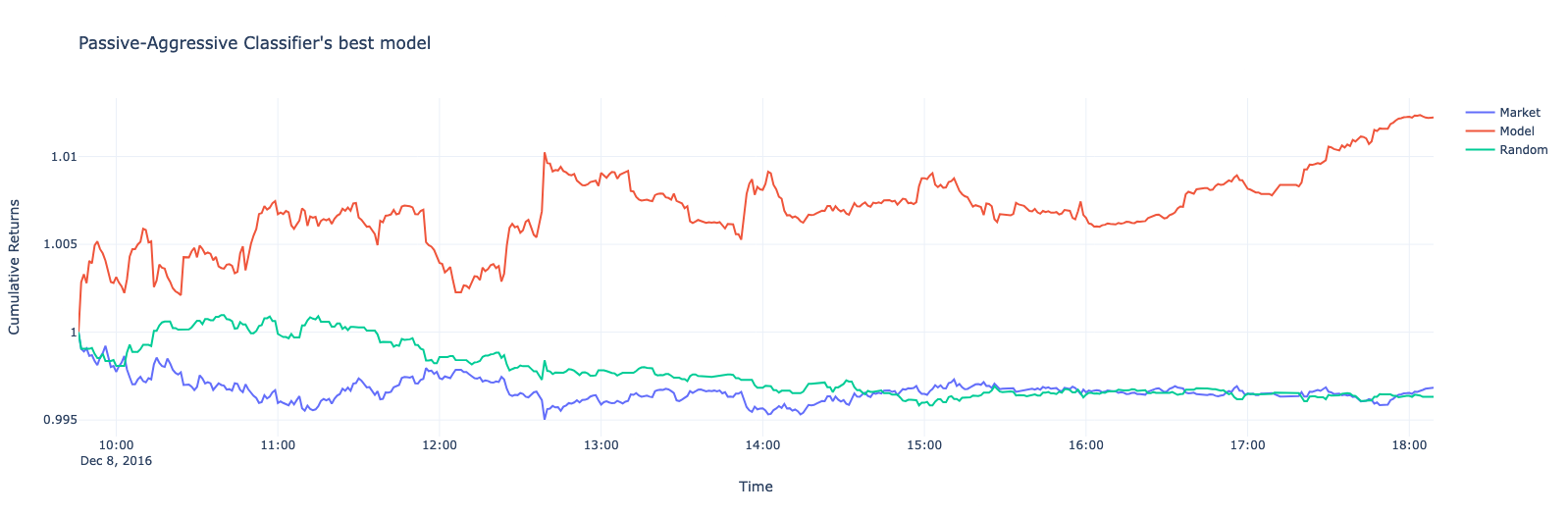}
\end{figure}

\section{3000 Observations Results}
\label{appendix:3000obs}
\begin{figure}[H]
    \centering
    \includegraphics[width=1\linewidth]{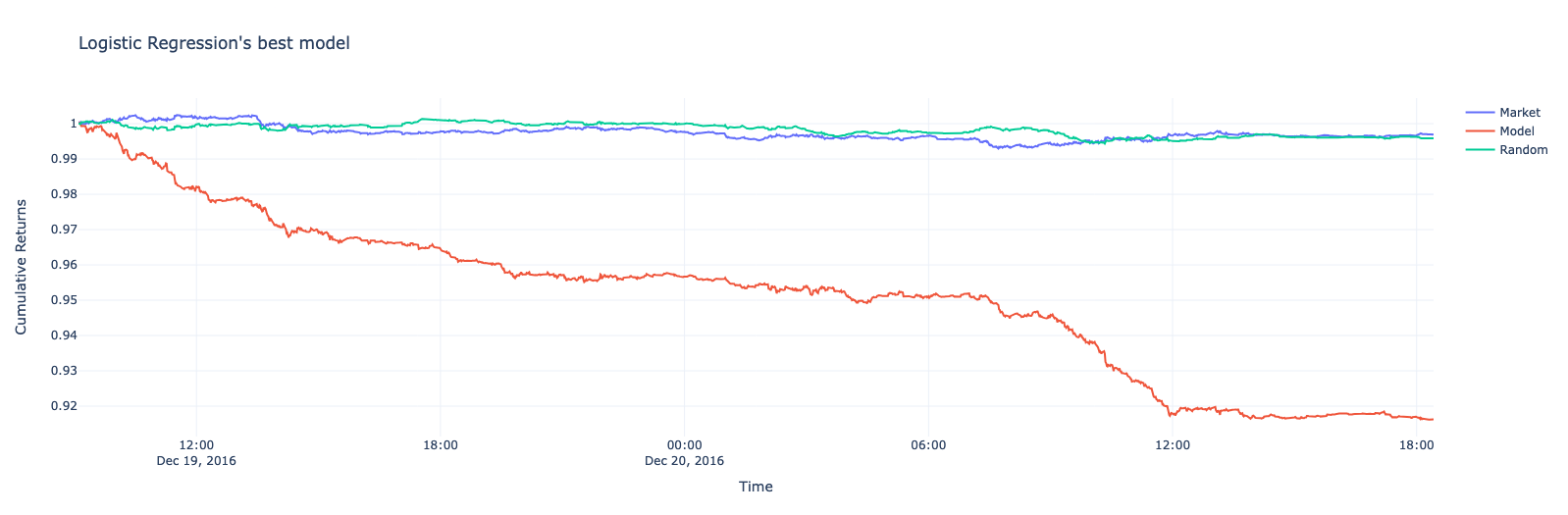}
\end{figure}
\begin{figure}[H]
    \centering
    \includegraphics[width=1\linewidth]{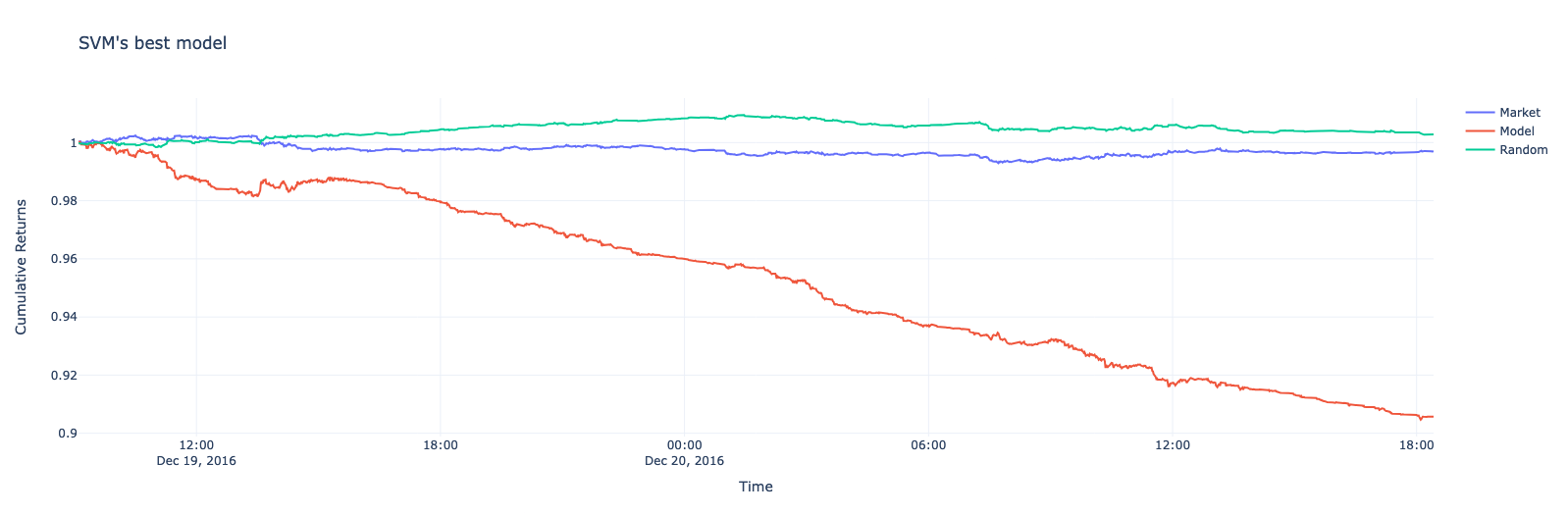}
\end{figure}
\begin{figure}[H]
    \centering
    \includegraphics[width=1\linewidth]{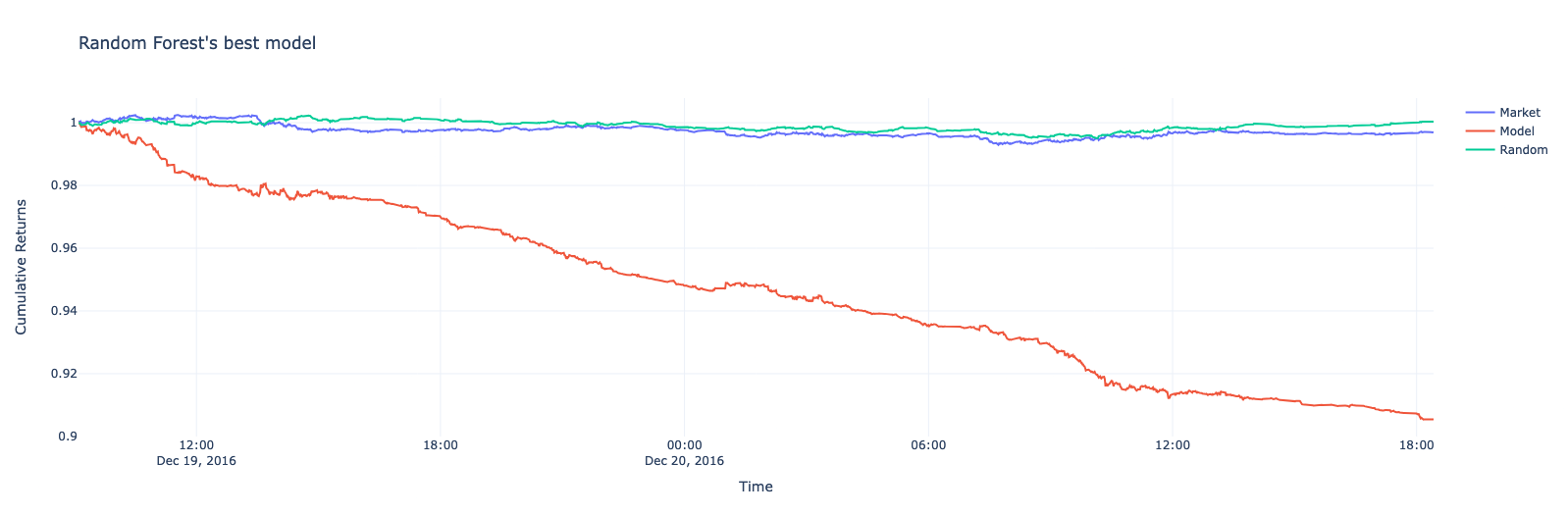}
\end{figure}
\begin{figure}[H]
    \centering
    \includegraphics[width=1\linewidth]{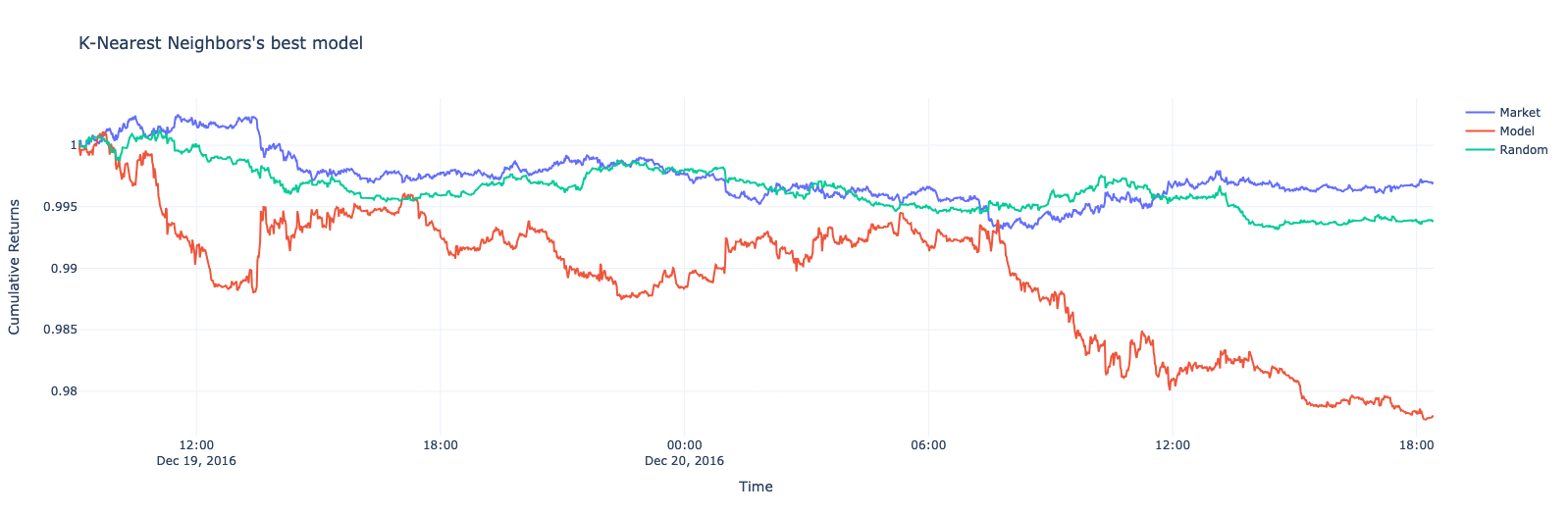}
\end{figure}
\begin{figure}[H]
    \centering
    \includegraphics[width=1\linewidth]{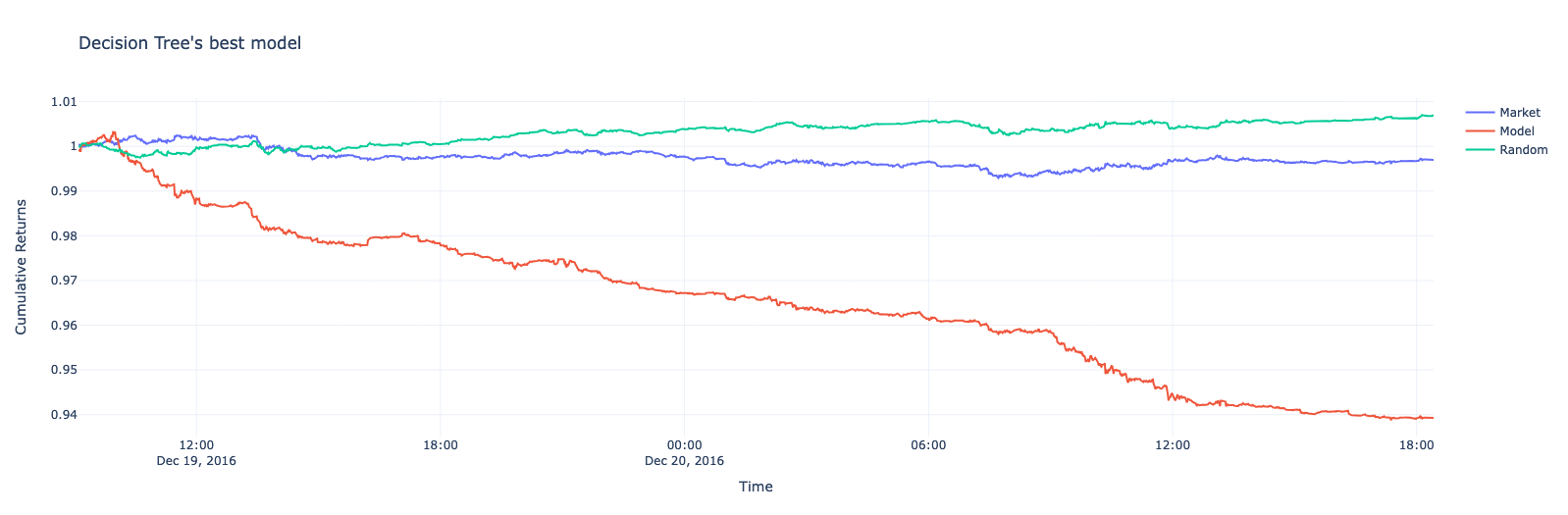}
\end{figure}
\begin{figure}[H]
    \centering
    \includegraphics[width=1\linewidth]{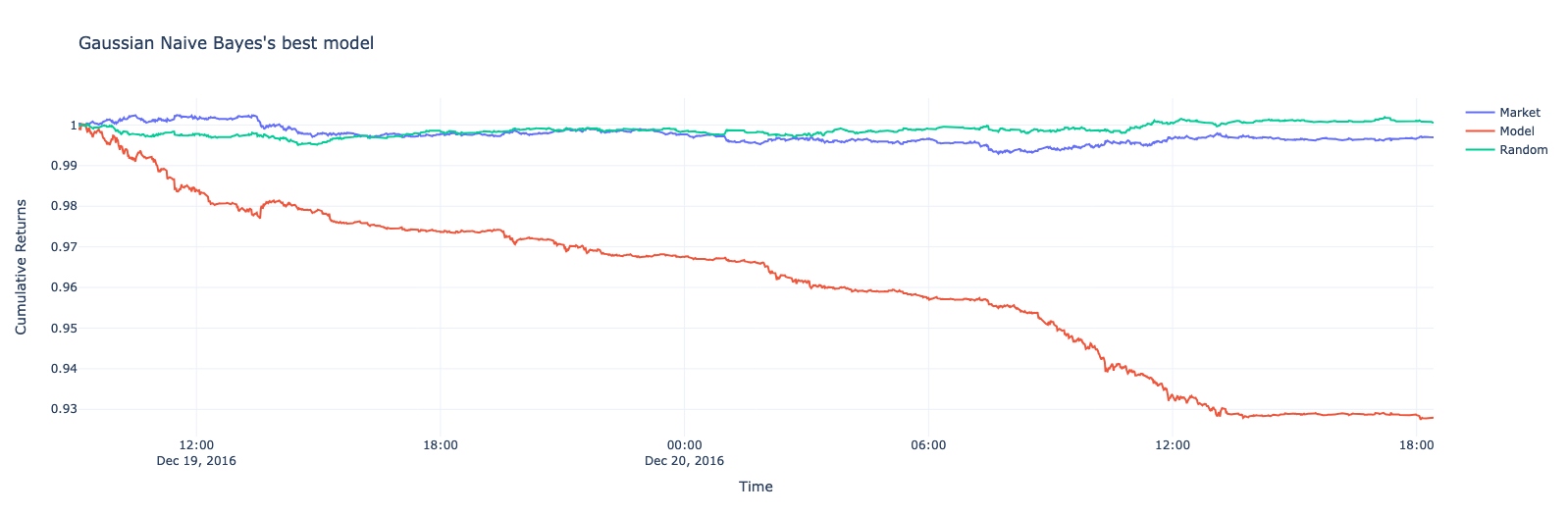}
\end{figure}
\begin{figure}[H]
    \centering
    \includegraphics[width=1\linewidth]{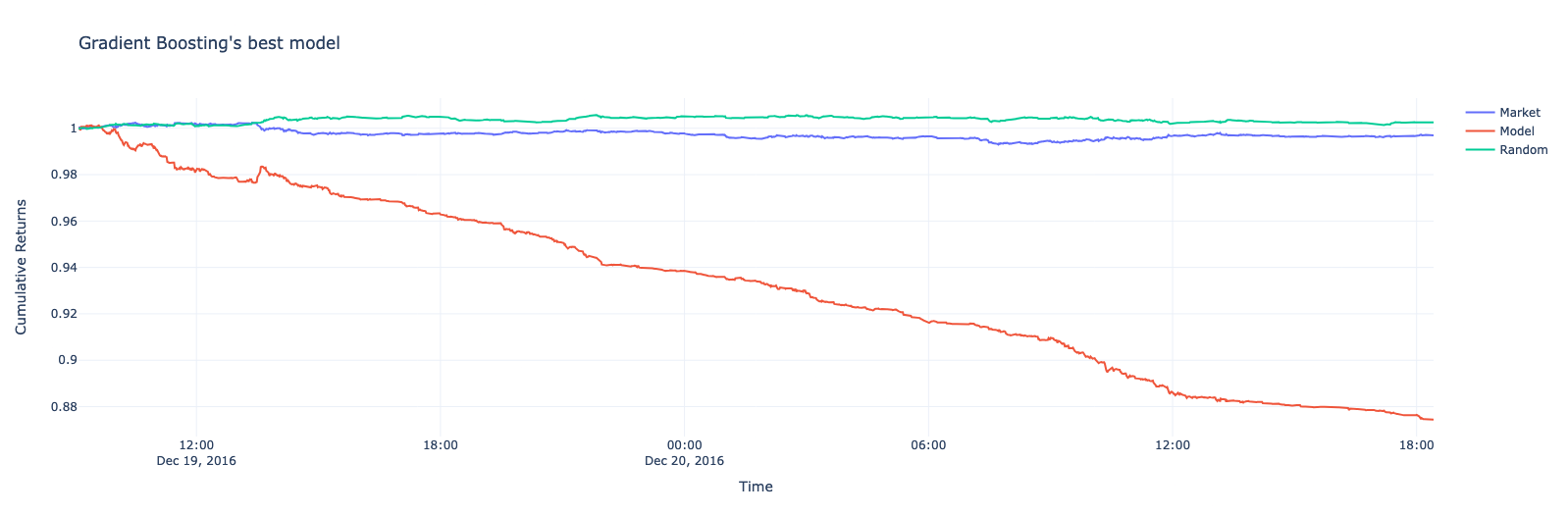}
\end{figure}
\begin{figure}[H]
    \centering
    \includegraphics[width=1\linewidth]{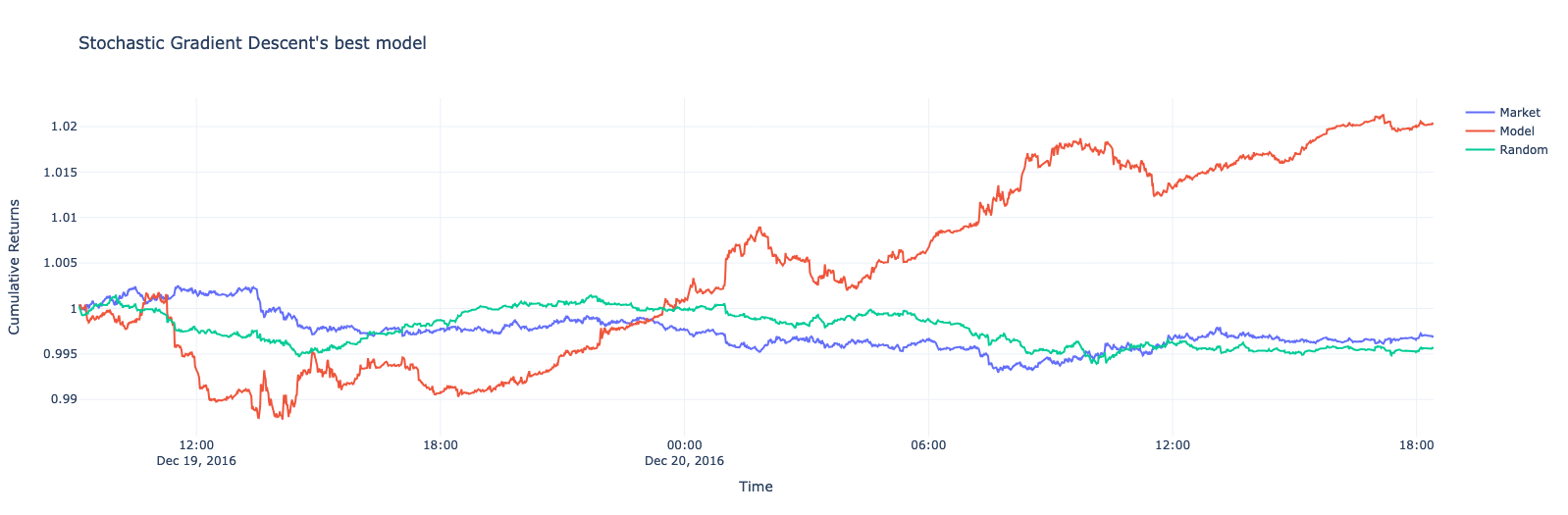}
\end{figure}
\begin{figure}[H]
    \centering
    \includegraphics[width=1\linewidth]{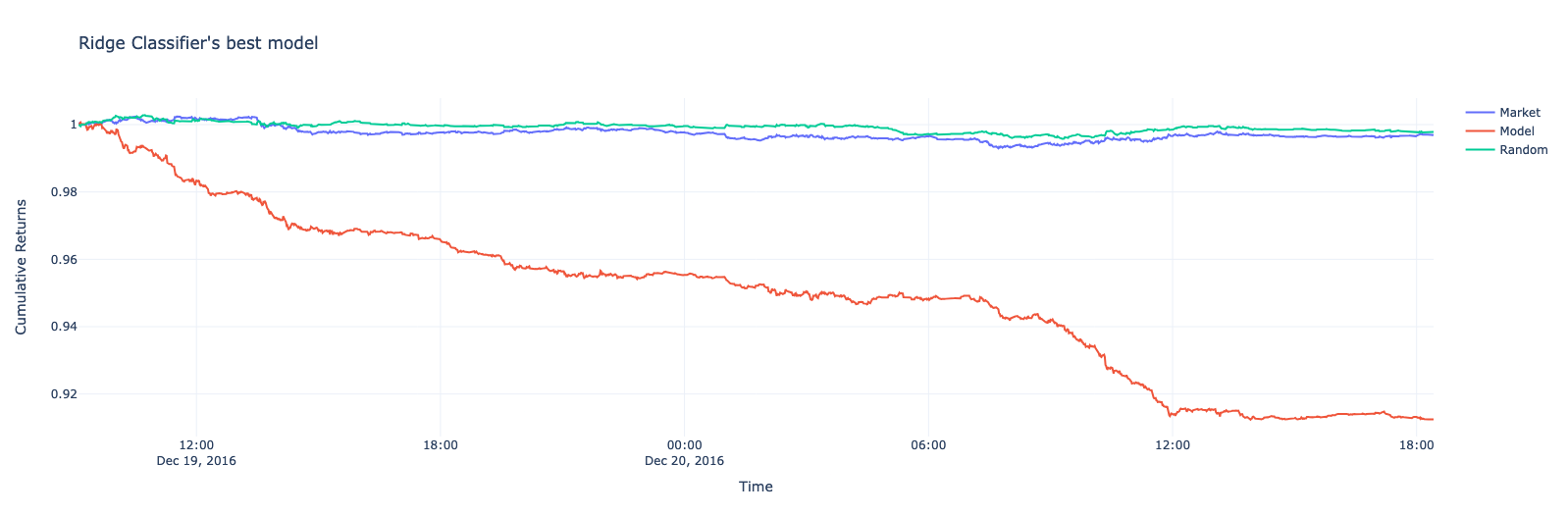}
\end{figure}
\begin{figure}[H]
    \centering
    \includegraphics[width=1\linewidth]{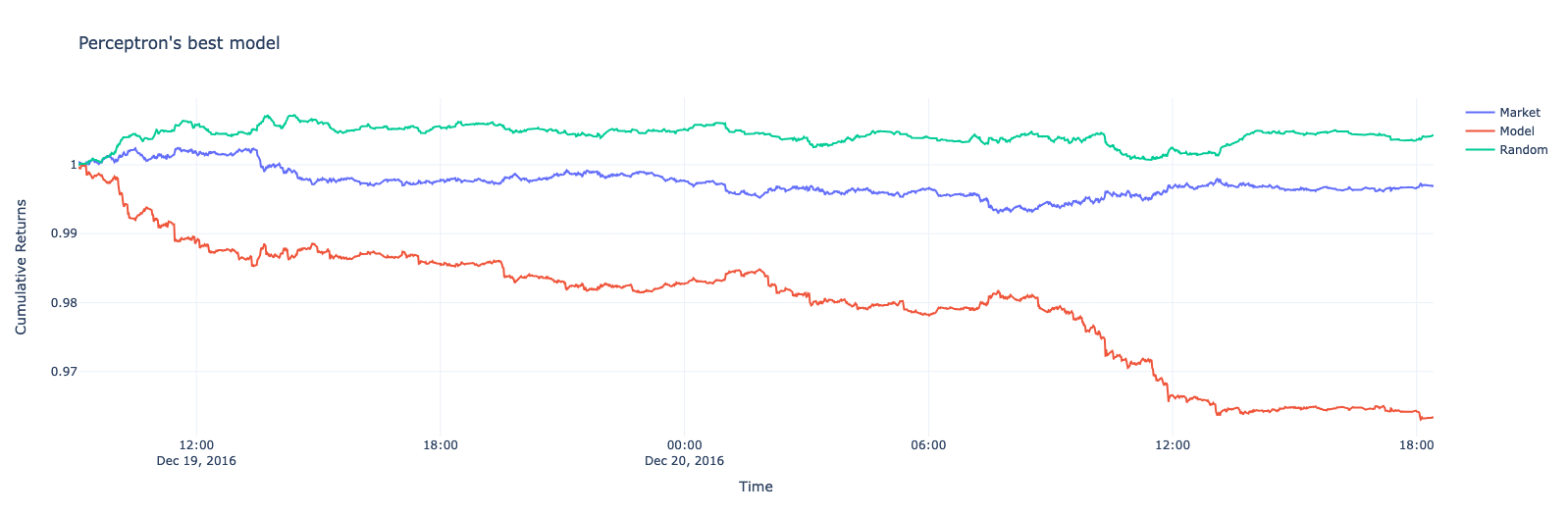}
\end{figure}
\begin{figure}[H]
    \centering
    \includegraphics[width=1\linewidth]{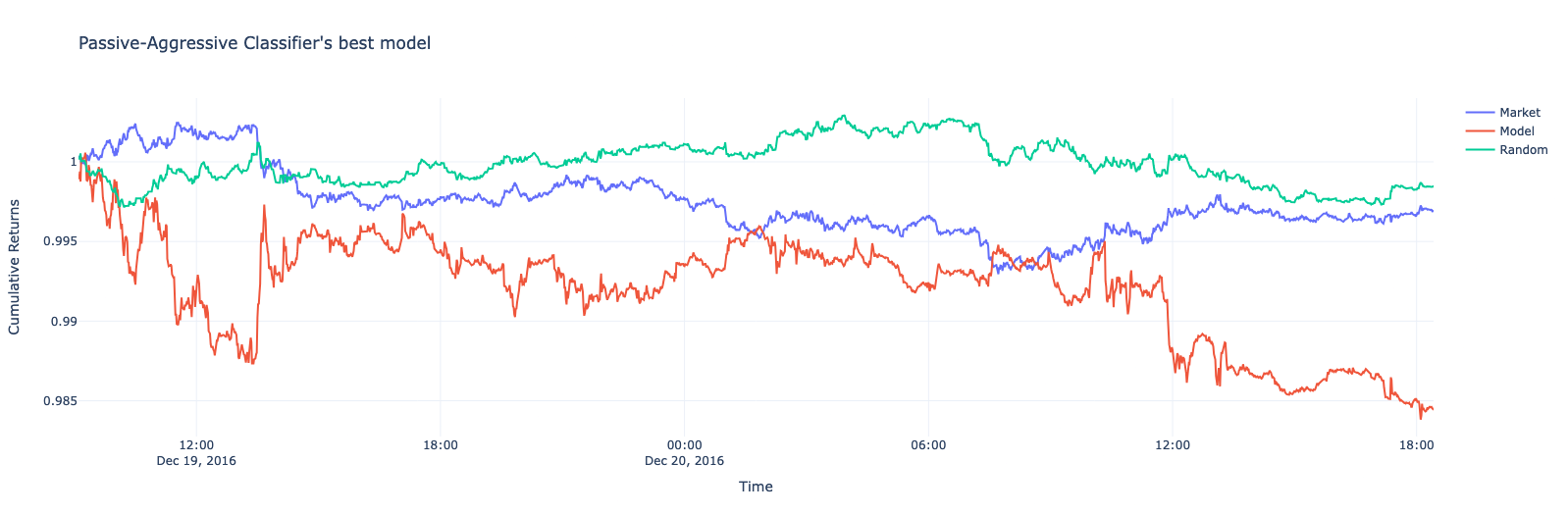}
\end{figure}

\section{\texttt{convTimeNet}}
\label{appendix:convTimeNet}
\begin{lstlisting}[language=Python]
from sklearn.preprocessing import StandardScaler
import torch
from torch.utils.data import DataLoader, TensorDataset
import pandas as pd
import numpy as np

from sklearn.preprocessing import StandardScaler
import torch
from torch.utils.data import DataLoader, TensorDataset
import pandas as pd
import numpy as np
import torch.nn as nn
import plotly.graph_objects as go

import warnings
from sklearn.exceptions import ConvergenceWarning

warnings.simplefilter(action='ignore', category=ConvergenceWarning)
warnings.simplefilter(action='ignore', category=pd.errors.PerformanceWarning)
warnings.simplefilter(action='ignore', category=pd.errors.SettingWithCopyWarning)

df_min = df.iloc[:100_000]

features = df_min.drop(['target'], axis=1)
target = df_min['target']

features = features.select_dtypes(include=[np.number])

scaler = StandardScaler()
features_scaled = scaler.fit_transform(features)

target_encoded = target.values + 1

features_tensor = torch.tensor(features_scaled, dtype=torch.float32)
target_tensor = torch.tensor(target_encoded, dtype=torch.long)

dataset = TensorDataset(features_tensor, target_tensor)
dataloader = DataLoader(dataset, batch_size=32, shuffle=True)

class ConvTimeNet(nn.Module):
    def __init__(self, num_features, num_classes):
        super(ConvTimeNet, self).__init__()
        self.conv1 = nn.Conv1d(in_channels=num_features, out_channels=64, kernel_size=3, padding=1)
        self.bn1 = nn.BatchNorm1d(64)
        self.relu = nn.ReLU()
        self.conv2 = nn.Conv1d(in_channels=64, out_channels=128, kernel_size=3, padding=1)
        self.bn2 = nn.BatchNorm1d(128)
        self.pool = nn.AdaptiveAvgPool1d(output_size=1)
        self.flatten = nn.Flatten()
        self.dropout = nn.Dropout(0.5)
        self.fc = nn.Linear(128, num_classes)

    def forward(self, x):
        x = self.relu(self.bn1(self.conv1(x)))
        x = self.relu(self.bn2(self.conv2(x)))
        x = self.pool(x)
        x = self.flatten(x)
        x = self.dropout(x)
        x = self.fc(x)
        return x


model = ConvTimeNet(num_features=features_tensor.shape[1], num_classes=3)
criterion = nn.CrossEntropyLoss()
optimizer = torch.optim.Adam(model.parameters(), lr=0.00001)

epochs = 100
for epoch in range(epochs):
    model.train()
    for inputs, labels in dataloader:
        optimizer.zero_grad()
        inputs = inputs.unsqueeze(2)
        outputs = model(inputs)
        loss = criterion(outputs, labels)
        loss.backward()
        optimizer.step()
    print(f'Epoch {epoch+1}, Loss: {loss.item()}')

def predict(model, dataloader):
    model.eval()
    predictions = []
    with torch.no_grad():
        for inputs, _ in dataloader:
            inputs = inputs.unsqueeze(2)
            outputs = model(inputs)
            _, predicted = torch.max(outputs, 1)
            predicted = predicted.numpy() - 1
            predictions.extend(predicted)
    return np.array(predictions)

predicted_signals = predict(model, dataloader)
df_min['predicted_signals'] = predicted_signals

df_min['strategy_returns'] = df_min['returns'] * df_min['predicted_signals']

random_strategy = np.random.choice([-1, 0, 1], size=len(df_min))
df_min['random_strategy_returns'] = df_min['returns'] * random_strategy

cumulative_returns = {
    'Market': (1 + df_min['returns']).cumprod(),
    'Model': (1 + df_min['strategy_returns']).cumprod(),
    'Random': (1 + df_min['random_strategy_returns']).cumprod()
}

fig = go.Figure()
for key, values in cumulative_returns.items():
    fig.add_trace(go.Scatter(x=df_min.index, y=values, mode='lines', name=f'{key}'))
fig.update_layout(title='Model vs Market vs Random Strategy Cumulative Returns',
                  xaxis_title='Time',xaxis_type='category', yaxis_title='Cumulative Returns',
                  template='plotly_white')
fig.show()

\end{lstlisting}

\pagebreak
\bibliographystyle{apalike}
\bibliography{bib.bib}

\begin{thebibliography}{}

\bibitem[Abe and Abe, 2010]{abe2010feature}
Abe, S. and Abe, S. (2010).
\newblock Feature selection and extraction.
\newblock {\em Support vector machines for pattern classification}, pages 331--341.

\bibitem[Asif et~al., 2021]{asif2021graph}
Asif, N.~A., Sarker, Y., Chakrabortty, R.~K., Ryan, M.~J., Ahamed, M.~H., Saha, D.~K., Badal, F.~R., Das, S.~K., Ali, M.~F., Moyeen, S.~I., et~al. (2021).
\newblock Graph neural network: A comprehensive review on non-euclidean space.
\newblock {\em Ieee Access}, 9:60588--60606.

\bibitem[Cheng et~al., 2024]{cheng2024convtimenet}
Cheng, M., Yang, J., Pan, T., Liu, Q., and Li, Z. (2024).
\newblock Convtimenet: A deep hierarchical fully convolutional model for multivariate time series analysis.
\newblock {\em arXiv preprint arXiv:2403.01493}.

\bibitem[Christ et~al., 2016]{christ2016distributed}
Christ, M., Kempa-Liehr, A.~W., and Feindt, M. (2016).
\newblock Distributed and parallel time series feature extraction for industrial big data applications.
\newblock {\em arXiv preprint arXiv:1610.07717}.

\bibitem[Cortes, 1995]{cortes1995support}
Cortes, C. (1995).
\newblock Support-vector networks.
\newblock {\em Machine Learning}.

\bibitem[Dixon et~al., 2017]{dixon2017classification}
Dixon, M., Klabjan, D., and Bang, J.~H. (2017).
\newblock Classification-based financial markets prediction using deep neural networks.
\newblock {\em Algorithmic Finance}, 6(3-4):67--77.

\bibitem[Engle and Ng, 1982]{engle1982introduction}
Engle, R.~F. and Ng, V. (1982).
\newblock An introduction to the use of arch/garch models in applied econometrics.
\newblock {\em Journal of Business, New York}.

\bibitem[Fons et~al., 2020]{fons2020evaluating}
Fons, E., Dawson, P., Zeng, X.-j., Keane, J., and Iosifidis, A. (2020).
\newblock Evaluating data augmentation for financial time series classification.
\newblock {\em arXiv preprint arXiv:2010.15111}.

\bibitem[Fukushima, 1969]{fukushima1969visual}
Fukushima, K. (1969).
\newblock Visual feature extraction by a multilayered network of analog threshold elements.
\newblock {\em IEEE Transactions on Systems Science and Cybernetics}, 5(4):322--333.

\bibitem[Gen{\c{c}}ay and Qi, 2001]{genccay2001pricing}
Gen{\c{c}}ay, R. and Qi, M. (2001).
\newblock Pricing and hedging derivative securities with neural networks: Bayesian regularization, early stopping, and bagging.
\newblock {\em IEEE transactions on neural networks}, 12(4):726--734.

\bibitem[Goldstein et~al., 2021]{goldstein2021big}
Goldstein, I., Spatt, C.~S., and Ye, M. (2021).
\newblock Big data in finance.
\newblock {\em The Review of Financial Studies}, 34(7):3213--3225.

\bibitem[He et~al., 2019]{he2019transfer}
He, Q.-Q., Pang, P. C.-I., and Si, Y.-W. (2019).
\newblock Transfer learning for financial time series forecasting.
\newblock In {\em PRICAI 2019: Trends in Artificial Intelligence: 16th Pacific Rim International Conference on Artificial Intelligence, Cuvu, Yanuca Island, Fiji, August 26--30, 2019, Proceedings, Part II 16}, pages 24--36. Springer.

\bibitem[Hearst et~al., 1998]{hearst1998support}
Hearst, M.~A., Dumais, S.~T., Osuna, E., Platt, J., and Scholkopf, B. (1998).
\newblock Support vector machines.
\newblock {\em IEEE Intelligent Systems and their applications}, 13(4):18--28.

\bibitem[Hochreiter, 1997]{hochreiter1997long}
Hochreiter, S. (1997).
\newblock Long short-term memory.
\newblock {\em Neural Computation MIT-Press}.

\bibitem[Ismail~Fawaz et~al., 2019]{ismail2019deep}
Ismail~Fawaz, H., Forestier, G., Weber, J., Idoumghar, L., and Muller, P.-A. (2019).
\newblock Deep learning for time series classification: a review.
\newblock {\em Data mining and knowledge discovery}, 33(4):917--963.

\bibitem[Krešňáková et~al., 2020]{vanishing}
Krešňáková, V., Sarnovsky, M., Butka, P., and Machova, K. (2020).
\newblock Comparison of deep learning models and various text pre-processing techniques for the toxic comments classification.
\newblock {\em Applied Sciences}, 10:8631.

\bibitem[Kumar et~al., 2013]{kumar2013verification}
Kumar, A., Kiran, M., and Prathap, B. (2013).
\newblock Verification and validation of mapreduce program model for parallel k-means algorithm on hadoop cluster.
\newblock In {\em 2013 Fourth International Conference on Computing, Communications and Networking Technologies (ICCCNT)}, pages 1--8. IEEE.

\bibitem[Kwak, 2013]{kwak2013nonlinear}
Kwak, N. (2013).
\newblock Nonlinear projection trick in kernel methods: An alternative to the kernel trick.
\newblock {\em IEEE transactions on neural networks and learning systems}, 24(12):2113--2119.

\bibitem[Lemley et~al., 2017]{transferfigure}
Lemley, J., Bazrafkan, S., and Corcoran, P. (2017).
\newblock Transfer learning of temporal information for driver action classification.

\bibitem[Lo, 2004]{lo2004adaptive}
Lo, A.~W. (2004).
\newblock The adaptive markets hypothesis: Market efficiency from an evolutionary perspective.
\newblock {\em Journal of Portfolio Management, Forthcoming}.

\bibitem[Lunga et~al., 2020]{lunga2020apache}
Lunga, D., Gerrand, J., Yang, L., Layton, C., and Stewart, R. (2020).
\newblock Apache spark accelerated deep learning inference for large scale satellite image analytics.
\newblock {\em IEEE Journal of Selected Topics in Applied Earth Observations and Remote Sensing}, 13:271--283.

\bibitem[Rumelhart et~al., 1986]{rumelhart1986learning}
Rumelhart, D.~E., Hinton, G.~E., and Williams, R.~J. (1986).
\newblock Learning internal representations by error propagation, parallel distributed processing, explorations in the microstructure of cognition, ed. de rumelhart and j. mcclelland. vol. 1. 1986.
\newblock {\em Biometrika}, 71(599-607):6.

\bibitem[Susto et~al., 2018]{susto2018time}
Susto, G.~A., Cenedese, A., and Terzi, M. (2018).
\newblock Time-series classification methods: Review and applications to power systems data.
\newblock {\em Big data application in power systems}, pages 179--220.

\bibitem[Wang et~al., 2016]{wang2016attention}
Wang, Y., Huang, M., Zhu, X., and Zhao, L. (2016).
\newblock Attention-based lstm for aspect-level sentiment classification.
\newblock In {\em Proceedings of the 2016 conference on empirical methods in natural language processing}, pages 606--615.

\bibitem[Wu et~al., 2020]{wu2020labeling}
Wu, D., Wang, X., Su, J., Tang, B., and Wu, S. (2020).
\newblock A labeling method for financial time series prediction based on trends.
\newblock {\em Entropy}, 22(10):1162.

\bibitem[Yin et~al., 2019]{yin2019experimental}
Yin, J., Rao, W., Yuan, M., Zeng, J., Zhao, K., Zhang, C., Li, J., and Zhao, Q. (2019).
\newblock Experimental study of multivariate time series forecasting models.
\newblock In {\em Proceedings of the 28th ACM international conference on information and knowledge management}, pages 2833--2839.

\end{thebibliography}
%%%%%%%%%%%%%%%%%%%%%%%%%%%%%%%%%%%%%%%%%%%%%%%%%%%%%

\end{document}